\documentclass[lettersize,journal]{IEEEtran}
\usepackage{amsmath,amsfonts}
\usepackage{algorithmic}
\usepackage{algorithm}
\usepackage{array}
\usepackage{textcomp}
\usepackage{stfloats}
\usepackage{url}
\usepackage{xspace}
\usepackage{verbatim}
\usepackage{graphicx}
\usepackage{cite}
\usepackage{amsmath,graphicx,multirow,bbding,booktabs,amssymb,bm,bbm}
\usepackage{graphicx}
\usepackage{float}
\usepackage{subfigure}

\usepackage[pagebackref=true,breaklinks=true,letterpaper=true,colorlinks,citecolor=blue,bookmarks=false]{hyperref}

\def\OurDataset{{WuhanMetroCrowd}\xspace}

\begin{document}

\title{Crowded Video Individual Counting Informed by Social Grouping and Spatial-Temporal Displacement Priors}

\author{Hao Lu,~\IEEEmembership{Senior Member,~IEEE}, Xuhui Zhu, Wenjing Zhang, Yanan Li,~\IEEEmembership{Senior Member,~IEEE},\\
Xiang Bai,~\IEEEmembership{Fellow,~IEEE}
\thanks{This work is supported by the National Natural Science Foundation of China under Grant No.~62576146, and in part by Hubei Provincial Colleges and Universities Outstanding Young and Middle-aged Science under Grant T2023009, in part by the International Science and Technology Cooperation Program of Hubei Province under Grant 2025EHA017, and in part by the Science and Technology Research Project of Hubei Provincial Department of Education under Grant B2024046. \textit{(Corresponding author: Yanan Li.)}}

\thanks{H. Lu, X. Zhu, and W. Zhang are with State Key Laboratory of Multispectral Information Intelligent Processing Technology; School of Artificial Intelligence and Automation, Huazhong University of Science and Technology, Wuhan 430074, China. (e-mail: \{hlu,xuhuizhu\}@hust.edu.cn, wenjingzhang1412@gmail.com)}

\thanks{Y. Li is with Hubei Key Laboratory of Intelligent Robot; School of Computer Science \& Engineering Artificial Intelligence, Wuhan Institute of Technology, Wuhan 430205, China. (e-mail: yananli@wit.edu.cn)}

\thanks{X. Bai is with School of Software Engineering, Huazhong University of Science and Technology, Wuhan 430074, China. (e-mail: xbai@hust.edu.cn)}

}

\markboth{Manuscript submitted to IEEE Trans. Image Process.; V1 Jan, 2026.
}%
{Lu \MakeLowercase{\textit{et al.}}: Crowded Video Individual Counting}


\maketitle

\begin{abstract}
Video Individual Counting (VIC) is a recently introduced task aiming to estimate pedestrian flux from a video. 
It extends Video Crowd Counting (VCC) beyond the per-frame pedestrian count.
In contrast to VCC that learns to count pedestrians across frames, VIC must identify co-existent pedestrians between frames, which turns out to be a correspondence problem. 
Existing VIC approaches, however, can underperform in congested scenes such as metro commuting. 
To address this, we build \OurDataset, one of the first VIC datasets that characterize crowded, dynamic pedestrian flows.  
It features sparse-to-dense density levels, short-to-long video clips, slow-to-fast flow variations, front-to-back appearance changes, and light-to-heavy occlusions.
To better adapt VIC approaches to crowds, we rethink the nature of VIC and recognize two informative priors: \romannumeral1) the social grouping prior that indicates pedestrians tend to gather in groups and \romannumeral2) the spatial-temporal displacement prior that informs an individual cannot teleport physically.
The former inspires us to relax the standard one-to-one (O2O) matching used by VIC to one-to-many (O2M) matching, implemented by an implicit context generator and a O2M matcher; 
the latter facilitates the design of a displacement prior injector, which strengthens not only O2M matching but also feature extraction and model training.
These designs jointly form a novel and strong VIC baseline OMAN++.
Extensive experiments show that OMAN++ not only outperforms state-of-the-art VIC baselines on the standard SenseCrowd, CroHD, and MovingDroneCrowd benchmarks, but also indicates a clear advantage in crowded scenes, with a $38.12\%$ error reduction on our \OurDataset dataset. 
Code, data, and pretrained models are available at \url{https://github.com/tiny-smart/OMAN}.
\end{abstract}


\section{Introduction}

\IEEEPARstart{M}{odern} urbanization and population growth have led to increasingly crowded public spaces, raising concerns about public security. 
Metro systems, as a representative, face significant challenges in managing dense passenger flows, especially during commuting hours~\cite{metro_sys}. 
Effective and efficient crowd analysis is therefore essential for optimizing traffic flows and enabling crowd management. 
While video surveillance and vision-based techniques such as Video Crowd Counting (VCC)~\cite{LSTN,FTAN,PET} have been widely adopted to monitor crowds, VCC only addresses per-frame pedestrian counting, revealing deficiency in scenarios where the information of pedestrian flux within a period of time matters.
To address this, cross-line crowd counting, a.k.a. Line of Interest (LOI)~\cite{LOI-2013,LOI-2016,LOI-2019}, and Multiple Object Tracking (MOT)~\cite{ByteTrack,apptracker+,TransMOT} are introduced or repurposed to acquire such information. Yet, LOI cannot capture pedestrians of different walking directions, and the computational cost of MOT can increase quadratically with the number of pedestrians detected.

\label{sec:intro}
\begin{figure}[!t]
    \centering
    \vspace{-5pt}
    \centerline{\includegraphics[width=1\linewidth]{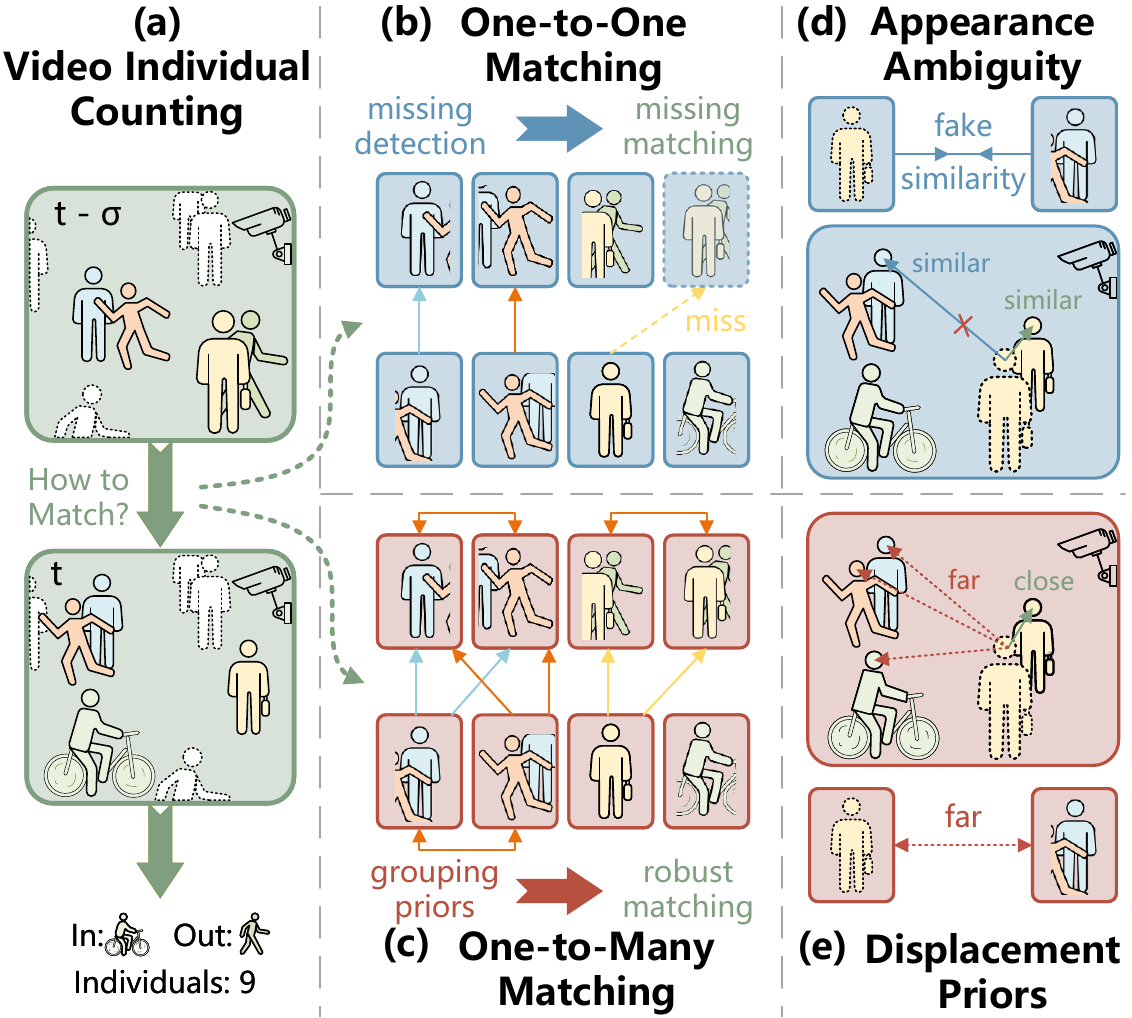}}
    \caption{ \textbf{Video individual counting in crowded scenes}. 
    (a) The key of VIC is to identify co-existent pedestrians between frames,
    but (b) existing methods typically apply a O2O matching strategy, suffering from missing detections due to occlusions. 
    (c) Inspired by the grouping prior of walking pedestrians, we relax the O2O matching to O2M matching that allows an individual to match a group to enhance matching robustness. 
    In addition, (d) matching informed by only appearance cues may not distinguish visually similar pedestrians, leading to wrong matches. 
    In fact, (e) walking pedestrians obey spatial-temporal displacement prior that encourage  spatially close matches.
    }
\label{fig:VIC}
\end{figure}

To better estimate pedestrian flux beyond LOI and MOT, a new task termed Video Individual Counting (VIC) is introduced recently~\cite{DRNet}. As shown in Fig.~\ref{fig:VIC}(a), 
VIC aims to count unique pedestrians across video frames to infer inflow/outflow numbers. 
While VIC still concerns counting, the essence of VIC turns out to be a correspondence problem, i.e., matching co-existent pedestrians between frames.
Following this paradigm, a number of VIC approaches have been developed with increasingly improved performance~\cite{DRNet,CGNet,PDTR,MDC};
however, we observe that they often underperform in crowded scenes. 
Such phenomena may stem from two facts: i) the deficiency of crowded VIC data and ii) the exact O2O matching conditioned on appearance only.
Crowded scenes, which are common in VCC datasets, are less presented in VIC datasets. 
A main reason seems that early VIC datasets are repurposed from MOT datasets~\cite{SenseCrowd,HeadHunter-T} such that the model does not see sufficient crowd patterns during training. 
In addition, crowded scenes typically characterize frequent occlusions and reduced appearance cues. 
The former can render wrong or missing pedestrian detections (Fig.~\ref{fig:VIC}(b)); 
the latter can raise great challenges for exact pedestrian matching, leading to wrong or missing matches (Fig.~\ref{fig:VIC}(d)).

To overcome the data dilemma, we introduce \OurDataset, one of the first VIC datasets tailored to crowded scenes based on the Wuhan metro system.
In particular, \OurDataset hosts $80$ surveillance videos from $15$ metro stations collected during peak hours, holidays, and festivals, spanning from $2023$ to $2025$. As a result, it exhibits visual challenges including sparse-to-dense density levels, short-to-long video clips, slow-to-fast flow variations, front-to-back appearance changes, and light-to-heavy occlusions. Comprehensive manual annotations ($223,662$ labels) are also provided. 

\begin{figure}[!t]
	\centering  
	\subfigbottomskip=1pt 
	\subfigcapskip=-3pt 
	\subfigure[Social Grouping Prior]{
		\includegraphics[width=0.98\linewidth]{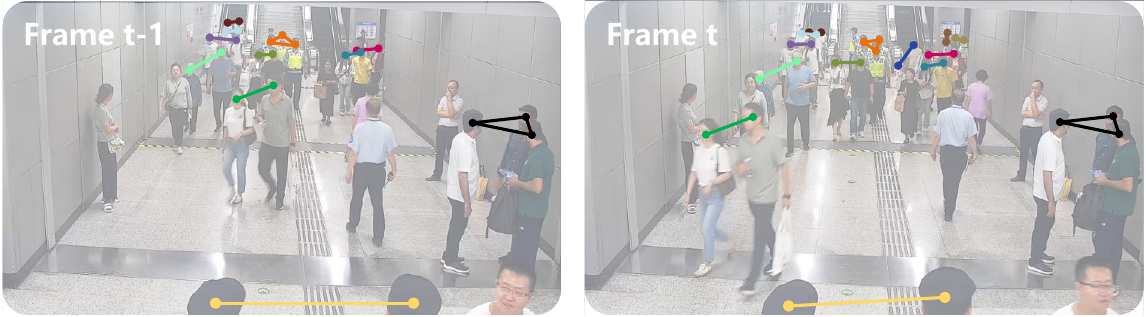}}
	\subfigure[Spatial-Temporal Displacement Prior]{
		\includegraphics[width=0.98\linewidth]{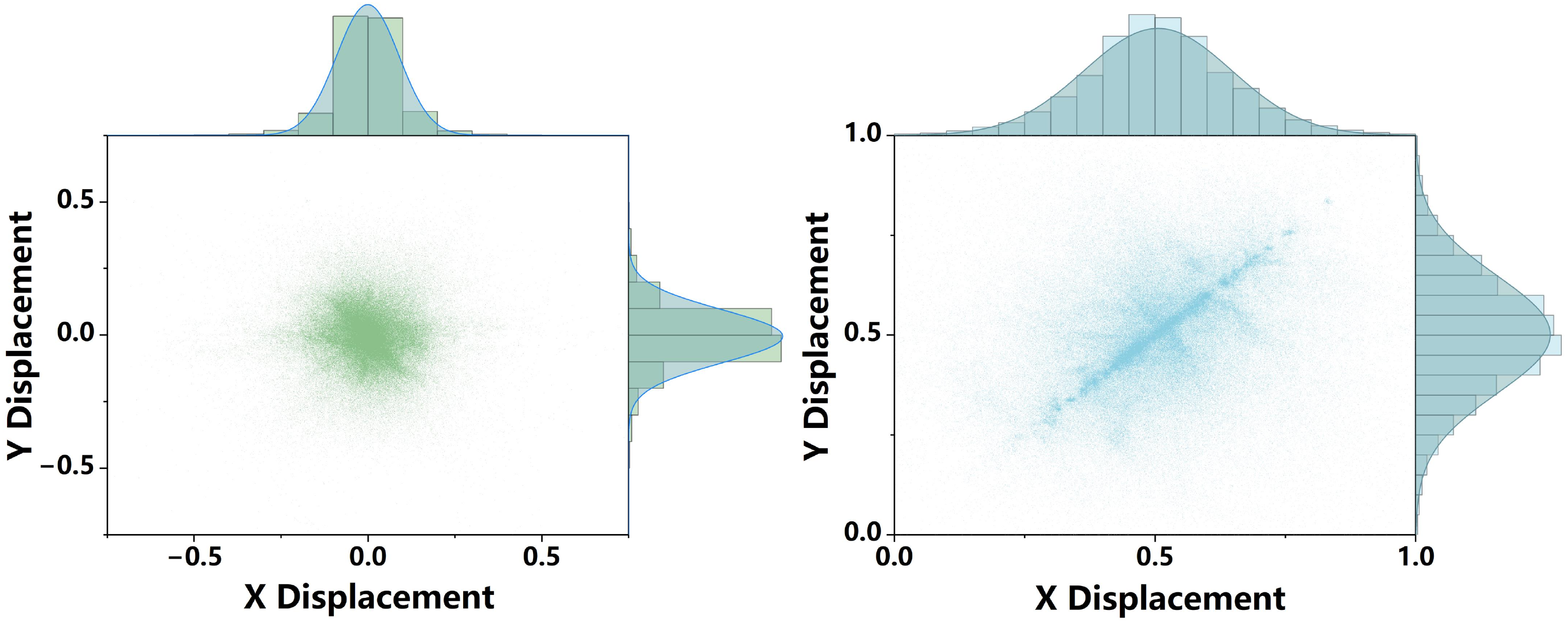}}
	\caption{\textbf{Two priors behind moving crowds}. (a) The social grouping prior indicating that pedestrians tend to walk in groups, with each social group connected with lines of different colors. (b) The spatial-temporal displacement prior revealing a limited range of pedestrian displacements (left) and their min-max normalized displacements (right) within three-second intervals, 
    suggesting pedestrian motion is physically bounded.} 
\label{fig:priors}
\end{figure}

To alleviate the technical challenges in pedestrian matching, we rethink the nature of VIC and have made two important observations: i) per Fig.~\ref{fig:priors}(a), pedestrians tend to gather by group, exhibiting a social grouping behavior; 
and ii) per Fig.~\ref{fig:priors}(b), pedestrians probably move within certain ranges between consecutive frames. 
The former suggests a social grouping prior that encourages the use of \textit{group context}; 
the latter implies a spatial-temporal displacement prior which reveals the fact that pedestrians cannot teleport physically and that pedestrian matching could be bounded locally. 



In this work, we develop our approach informed by the two priors above. 
We first propose to relax the O2O matching into O2M matching to harness the group context. 
In contrast to O2O matching that penalizes all wrong matches (Fig.~\ref{fig:VIC}(b)), O2M matching can leverage shared appearance and context cues to enable individual-to-group assignments (Fig.~\ref{fig:VIC}(c)), which alleviates occlusions. 
Indeed the nature of the VIC problem tells us that \textit{it is not necessary to match exactly the same pedestrian}. 
To exploit the displacement prior, we propose to incorporate it into the model to regularize both pedestrian matching and model training.

The O2M matching is implemented in a coarse-to-fine manner. In particular, an Implicit Context Generator (ICG) is introduced to infuse the group context into pedestrian descriptors. 
A One-to-Many Pairwise Matcher (OMPM) is then used to confirm the 
individual-to-group correspondence. This leads to a naive implementation termed OMAN. 
We then extend OMAN to OMAN++ by introducing a Displacement Prior Injector (DPI) to capsulate the spatio-temporal displacement prior via three synergistic modules: 
i) Displacement-Aware Self-Attention (DASA), which identifies geometrically plausible matches; 
ii) Displacement Modulator, which refines feature similarity; and iii) Displacement-informed Optimal Transport (D-OT) loss, which regularizes O2M matching. 
They work jointly to embed kinematic knowledge into the model. 

Experiments on three public benchmarks, including SenseCrowd~\cite{SenseCrowd}, CroHD~\cite{HeadHunter-T} and MovingDroneCrowd~\cite{MDC}, and our \OurDataset dataset reveal that OMAN++ emerges to be a simple yet strong baseline for VIC. On the three public benchmarks, it outperforms state-of-the-art approaches by $10.5\% \sim 34.7\%$ in MAE, $2.9\% \sim 20.9\%$ in MSE, and $5.5\% \sim 26.2\%$ in WRAE. Remarkably, OMAN++ reduces the MAE, MSE, and WRAE by $47.5\%$, $63.6\%$ and $38.1\%$, respectively, on our \OurDataset dataset, implying that our approach is dedicatedly optimized for crowded scenes. 
Ablation studies have also justified the contribution of each design. 
Aside from these results, we further study some practical issues such as the influence of the frame interval and the strategies used for incorporating the displacement prior. 
Our results suggest that \textit{O2M matching is a more natural matching paradigm than O2O for VIC} and \textit{the displacement prior is a instrumental cue for crowded VIC}.

Our contributions include the following.
\begin{itemize}
    \item We suggest two informative VIC priors: the social grouping prior that describes grouped individuals and the spatial-temporal displacement prior that bounds inter-frame pedestrian motion.
    
    \item OMAN++: a simple and strong VIC baseline that implements O2M matching and motion-aware reasoning, which performs local individual-to-group matching by fusing appearance and motion cues, reducing the complexities of crowded VIC.
    
    \item \OurDataset: A novel VIC benchmark that highlights crowded scenes, comprises $80$ video sequences with $11,925$ frames, and features long sequences, large flow and scale variations, and significant occlusions.
\end{itemize}

A preliminary conference version of this work appeared in~\cite{OMAN}. We extend~\cite{OMAN} by \romannumeral1) building a novel VIC benchmark WuhanMetroCrowd with significant data diversity and scene variations, which fills the data gap of crowded scenes and long sequences in VIC; \romannumeral2) delving into the physical motion nature of VIC and identifying the spatial-temporal displacement prior; \romannumeral3) introducing the cross-frame displacement-aware self-attention and a displacement modulated O2M matcher; and \romannumeral4) designing a displacement prior loss to enhance the learning of both appearance and motion representations. Besides limited experiments in \cite{OMAN}, we now \romannumeral5) conduct comprehensive experiments on four different benchmarks with consistent and substantial improvements over the state of the art.

\section{Related Work}
To estimate pedestrian flux, early methods can be divided into two categories: \textit{video crowd counting} and \textit{multiple object tracking}; however, these methods struggle to provide precise inflow/outflow counts until a novel task called \textit{video individual counting} is introduced.

\subsection{Video Crowd Counting}
Existing VCC methods can be divided to into region-of-interest (ROI)-based VCC and line-of-interest (LOI)-based VCC. ROI-based methods aim to count per-frame pedestrians in videos. Early work~\cite{VCC-det1, VCC-det2} treats crowd counting as a detection problem by converting point-level labels into pseudo bounding boxes for model training, where pedestrian counts are determined by the number of bounding boxes. Due to the limited performance of detection models in crowded scenes, some recent work~\cite{VCC-det3, VCC-det4
} introduces local constraints, but suffers from inaccurate pseudo bounding boxes. Thus, most this line of methods shift toward density map-based counting ~\cite{VCC-dens1, 
VCC-dens4
}. Particularly, some work~\cite{LSTN, FTAN} focuses on exploiting spatial-temporal information to improve VCC. Despite the success in accurate counting, density map-based approaches cannot localize crowds. 
To balance between counting and localization, localization-based methods~\cite{VCC-loc1, PET, VCC-loc2
} are proposed recently, aiming to directly predict the center points of heads. 
Generally, ROI-based counting only infers per-frame pedestrian counts rather than pedestrian flux. 

LOI-based methods, a.k.a. cross-line crowd counting, count pedestrians who cross a pre-set line to estimate directed flux. Early work approaches the problem using different strategies: some work~\cite{LOI-2009, LOI-tip} accumulates blobs or capture velocities by motion models before regressing the crossing-line counts, while others rely on the detect-and-track framework~\cite{LOI-track1, LOI-track4}. Yet, these approaches either depend on motion segmentation or object detection, which struggles to tackle occlusions in crowds. To 
mitigate occlusions, temporal slices~\cite{LOI-2013, LOI-2016-tcsvt} are introduced to directly regress counts on the line. However, the slices discard much spatial information such that these slice-based methods do not generalize to dense scenes as well. 
To count dense crowds, \cite{LOI-2016, LOI-2019} decouples the task into ROI counting and velocity estimation so that local counts can be accumulated around the LOI per velocity. 
However, since it is impractical to define LOI for each scenario in reality, LOI-based methods fail to count pedestrians of different directions. 

In contrast to ROI-based methods that accumulate pedestrians in each frame and to LOI-based ones that only count those in certain directions, our methods directly estimate pedestrian flux across frames and over the entire image.

\subsection{Multiple Object Tracking}
Multiple Object Tracking (MOT), which simultaneously addresses localization, tracklet association, and re-identification (ReID), aims to track pedestrians with their identification (ID). Tracking by detection~\cite{track-tip, ByteTrack, Bot-sort, OCsort
} is the most widely adopted paradigm, which employs motion filters (e.g., Kalman filters) and IoU-based matching strategies to achieve detection and ReID. These methods, however, suffer from limited detection performance in crowded scenes and from reliance on high-frame-rate tracklet associations. 
Another alternative is detection and tracking, where IDs are directly associated by similarity-aware regression or attention. 
JDE~\cite{JDE} first integrates object detection and appearance association 
into a unified model, enabling real-time MOT. 
FairMOT~\cite{FairMOT} and ChainedTracker~\cite{chainedtracker} further 
generate detections and identity embedding jointly. 
Some other work~\cite{centertrack,
apptracker+
} directly operates on center points of heads to mitigate 
occlusions and 
scale variations. Recently, transformer-based trackers~\cite{trackformer, TransMOT, transtrack
} become dominance, implementing association by implicitly exploiting both spatial correspondence and temporal context. 

Though these methods seem to function as a solution to VIC, the frequent ID switches---due to poor quality of features for detection and association---can cause errors on flux estimation. The high computational cost can also hinder practical applications, because trackers usually operates frame by frame at high rates. 
Instead, our approach excludes identity recognition and infers frames at an interval of $2\sim 3$ seconds, avoiding potential ID mismatches and improving efficiency.

\subsection{Video Individual Counting}
In contrast to VCC and MOT, VIC estimates pedestrian flux in a distinct manner. 
DRNet~\cite{DRNet}, being the first work in this task, formulates VIC as a problem of pedestrian matching and predicts both pedestrian inflows and outflows. 
While it outperforms MOT approaches, it still follows the problem setup of MOT and uses the identity labels of individuals. 
With a deeper insight that VIC is about matching rather than tracking, CGNet~\cite{CGNet} replaces identity annotations with categorical ones and proposes a group-level matching loss for supervision. 
To execute matching, both DRNet~\cite{DRNet} and CGNet~\cite{CGNet} adopt the idea of explicit O2O matching such as Hungarian matching~\cite{Hungarian} and optimal transport~\cite{OT} to identify co-existent pedestrians between adjacent frames. 
In this way, empirical thresholds are often required to execute bipartite matching, leading to parameter sensitivity. 
Alternatively, FMDC~\cite{FMDC} and PDTR~\cite{PDTR} exploit regression to match the same pedestrian between frames, but they still rely on O2O correspondence to associate pedestrians.
More recently, MDC~\cite{MDC} integrates matching and counting using cross-frame cross-attention to search for homologous appearance. 
Notably, these regression-based methods rely on Gaussian-like density maps, rendering difficulty in establishing point-to-point correspondences between pedestrians. 

Instead of executing exact O2O matching, we propose to use O2M matching to facilitate the use of social grouping nature and contextual cues between pedestrians, improving robustness under occlusions and crowded scenes. 
To alleviate the reliance on varied appearance, we further exploit motion cues like displacement to inform our model with spatial-temporal context.

\begin{figure*}[!t]
  \centering
  \centerline{\includegraphics[width=18.3cm]{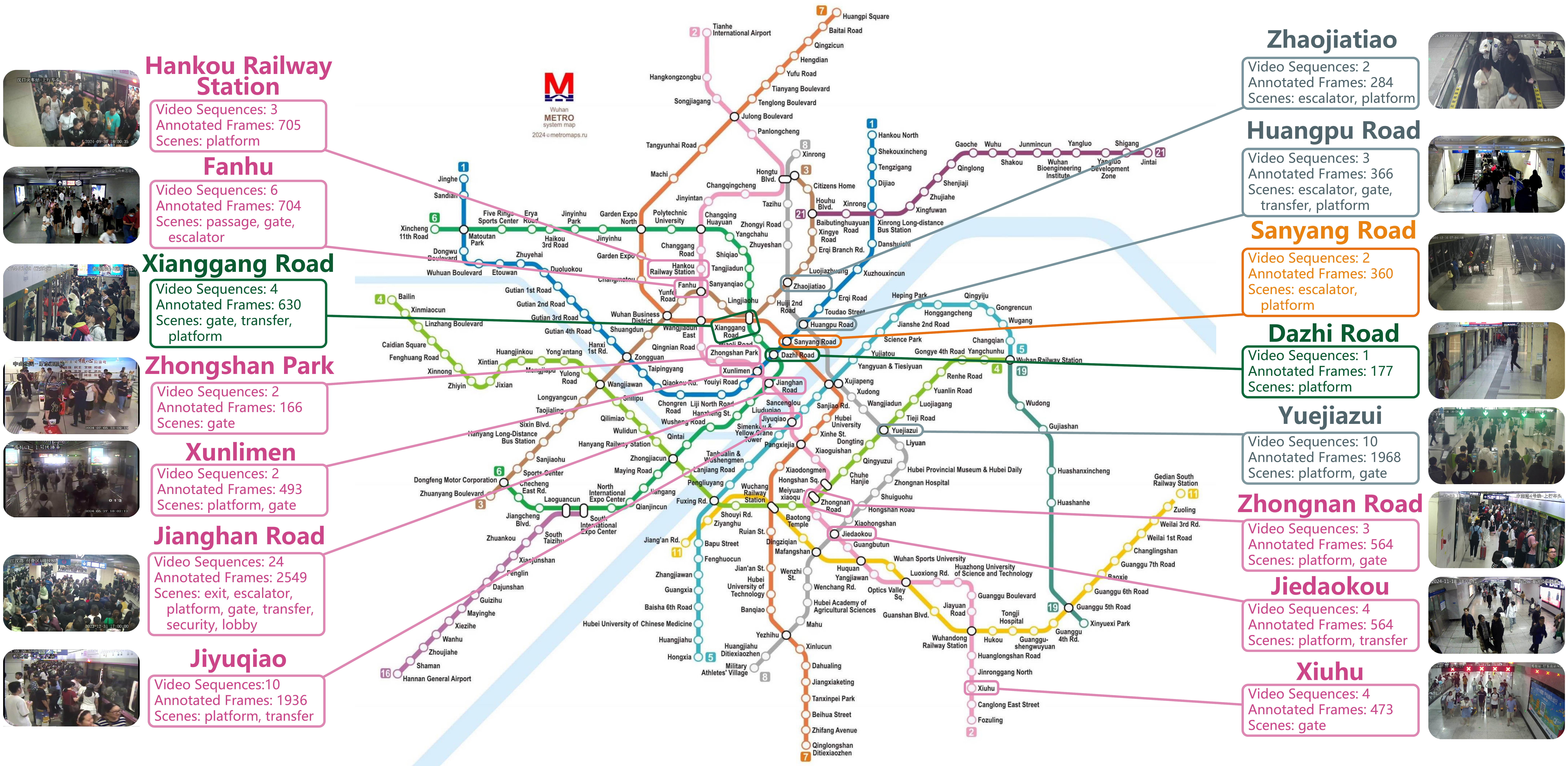}}
  \vspace{-5pt}
  \caption{\textbf{Overview of \OurDataset dataset}. This map shows the geographical distribution of data collection, and annotation number of each station.}
\label{fig:dataset overview}
\end{figure*}

\section{\OurDataset Dataset}
Existing VIC datasets primarily focus on short video clips with sparse or moderate crowd densities, where they fail to reflect the real-world challenges within long intervals and in crowded scenes, overlooking critical aspects such as accumulated errors, extreme density variations, severe occlusions, and large perspective changes.
To fill this gap, we introduce a novel, large-scale dataset, termed \OurDataset, tailored to dense VIC from a surveillance view, featuring diverse crowd dynamics and extended video clips. 
Particularly, \OurDataset highlights high-density scenarios, including densely packed crowds, persistent occlusions, and abrupt scale shifts, providing a thorough benchmark for evaluating VIC models in real-world scenarios. 
To our knowledge, \OurDataset is the first surveillance-perspective, crowded, and VIC-orientated dataset focusing on metro transport.
Details and properties of the dataset are illustrated in what follows.

\subsection{Data Collection, Annotation, and Partition}

\begin{figure*}[!t]
  \centering
  \centerline{\includegraphics[width=18cm]{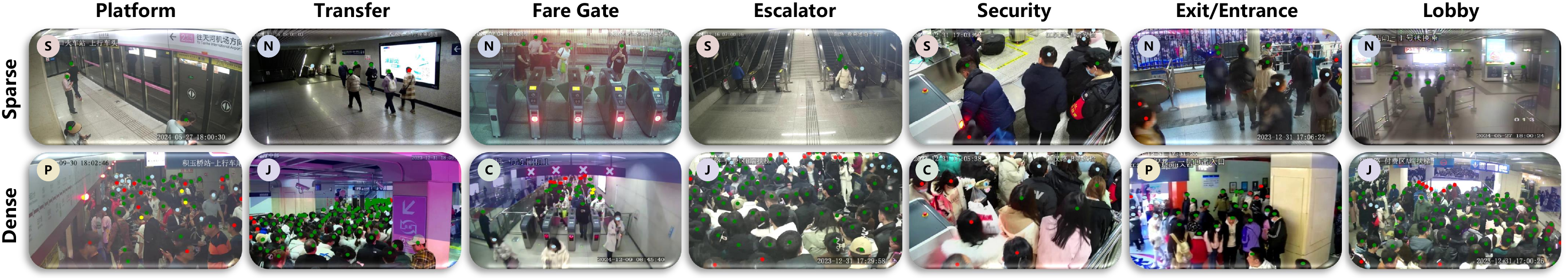}}
  \vspace{-5pt}
  \caption{\textbf{Annotated samples from \OurDataset dataset}. Scenes are shown on the top, and some relatively sparse scenes are shown in the first row while dense scenes in the second row. Green, red, light blue, and yellow points respectively represent the class `pedestrian', `inflow', `outflow', and `both', and transparent dark blue polygon regions represent masks. `S', `N', `C', `P', and `J' on the left-top of each picture, separately represents $5$ different density levels, `Sparse', `Normal', `Crowded', `Packed', and `Jam'.}
\label{fig:data_vis}
\end{figure*}

\begin{table*}[!t] \small
    \centering
    \caption{\textbf{Comparison with other datasets for VIC.} \textit{\#Frm./Seq.} represents frames per video sequence with $\sigma$-frame intervals, and \textit{\#Relative Flow Variation per Clip} represents the average of the flow variation values of people with different labels across all video clips. `-' is due to unreleased data.}
    \label{tab:dataset comparison}
    \renewcommand{\arraystretch}{1} 
    \addtolength{\tabcolsep}{-2.8pt} 
    \begin{tabular}{@{}lccccccccccc@{}}
        \toprule
        \multirow{2}{*}{Dataset} & \multirow{2}{*}{Task} & \multirow{2}{*}{Frames} & \multirow{2}{*}{Annotations} & \multirow{2}{*}{$\sigma$} & \multirow{2}{*}{Label} & \multirow{2}{*}{\#Frm./Seq.} & \multirow{2}{*}{View} & \multicolumn{4}{c}{\#Relative Flow Variation on Average}\\
        \cmidrule{9-12}
        &&&&&&&& Inflow & Outflow & Pedestrian & Total \\
        \midrule
        SenseCrowd~\cite{SenseCrowd} & MOT & 62,938 & 2,344,276 & 15 (3s) & ID & 6.62 & surveillance & 0.31 & 0.13 & 0.28 & 0.25\\
        CroHD~\cite{HeadHunter-T} & MOT & 11,464 & 2,275,286 & 75 (3s) & ID & 16.98 & drone & 0.03 & 0.03 & 0.46 & 0.46\\
        \midrule
        UAVVIC~\cite{CGNet} & VIC & 5,396 & 398,158 & 1 (3s) & In/Out & 24.42 & drone & - & - & - & -\\
        MovingDroneCrowd~\cite{MDC} & VIC & 4,940 & 325,542 & 4 (-) & ID & 13.88 & drone & 0.31 & 0.31 & 0.56 & 0.64\\
        \OurDataset (Ours) & VIC & 11,925 & 223,662 & 1 (2s) & In/Out & 149.06 & surveillance & 0.75 & 0.88 & 1.87 & 2.37\\
        \bottomrule
    \end{tabular}
\end{table*}

\begin{figure*}[!t]
	\centering  
	\subfigbottomskip=1pt 
	\subfigcapskip=-5pt 
	\subfigure[Statistics w.r.t. Sequences]{
		\includegraphics[width=0.24\linewidth,height=0.22\linewidth]{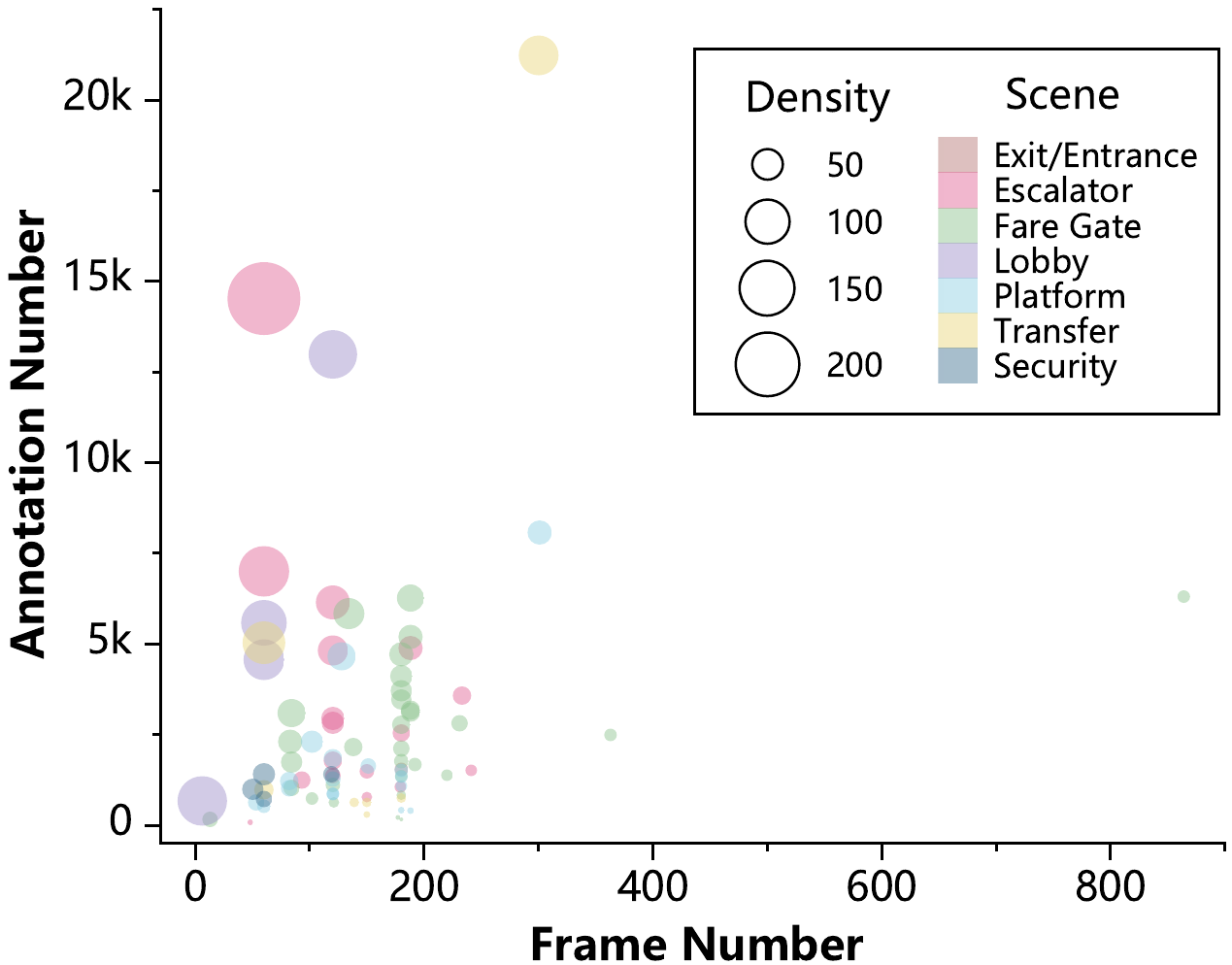}}
	\subfigure[Dataset Partition]{
		\includegraphics[width=0.23\linewidth]{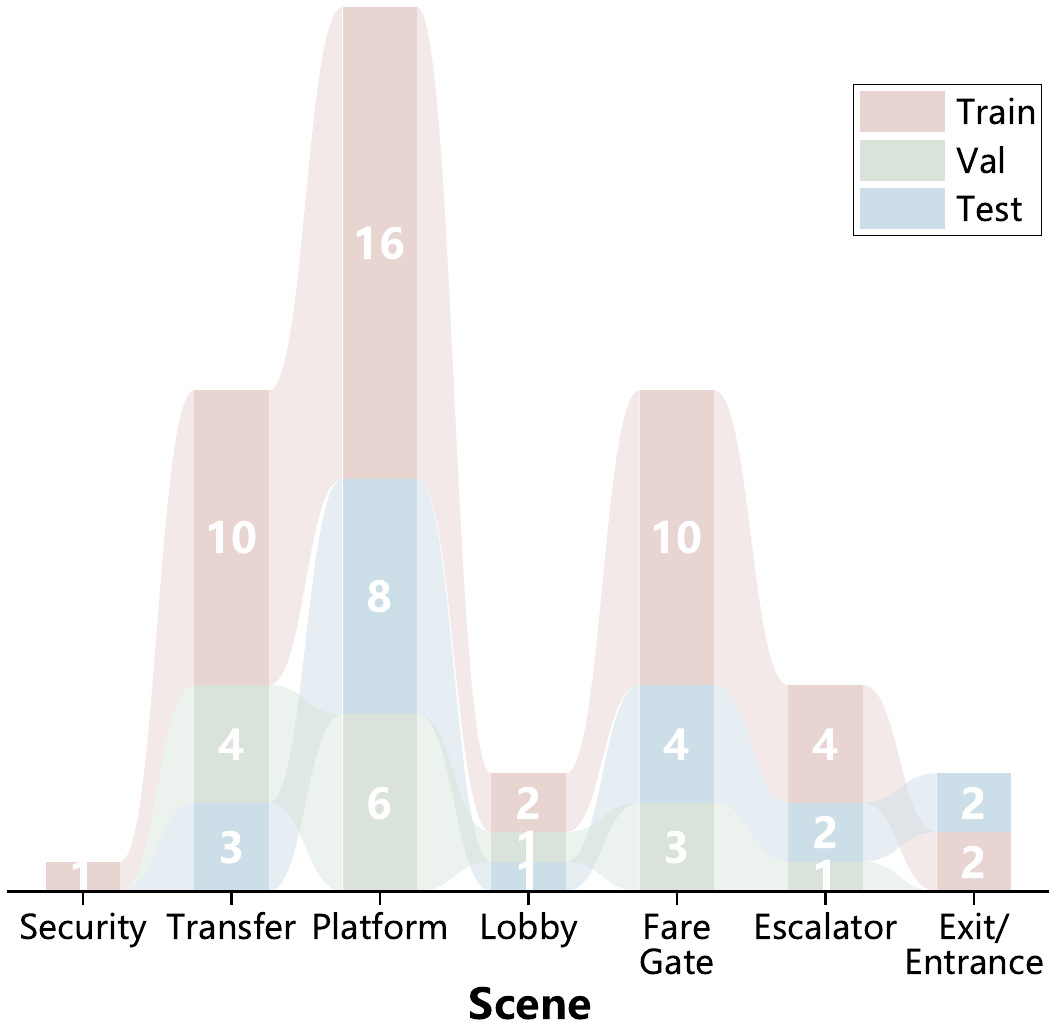}}
    \subfigure[Density Variation]{
		\includegraphics[width=0.24\linewidth]{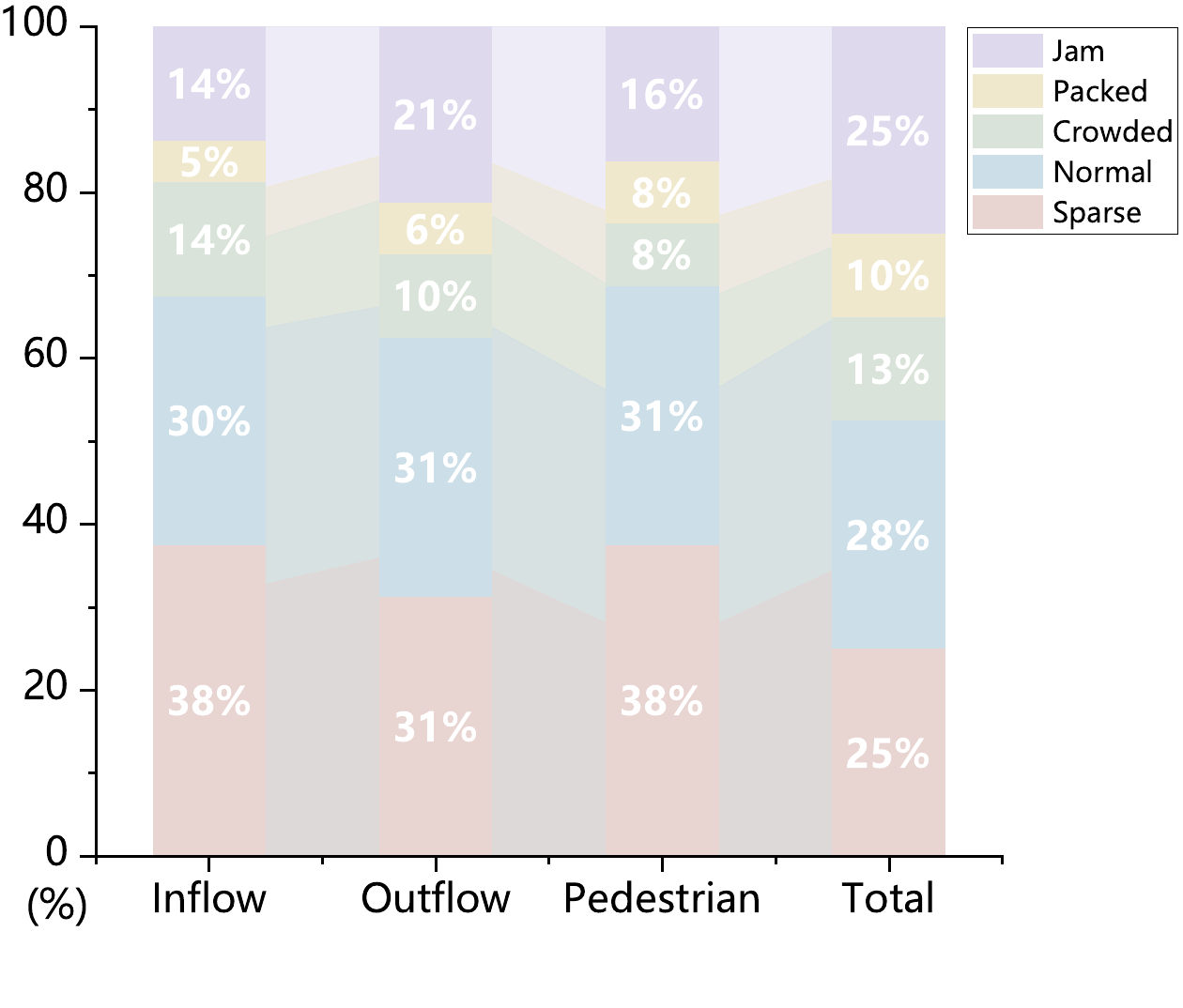}}
    \subfigure[Flow Variation]{
		\includegraphics[width=0.24\linewidth]{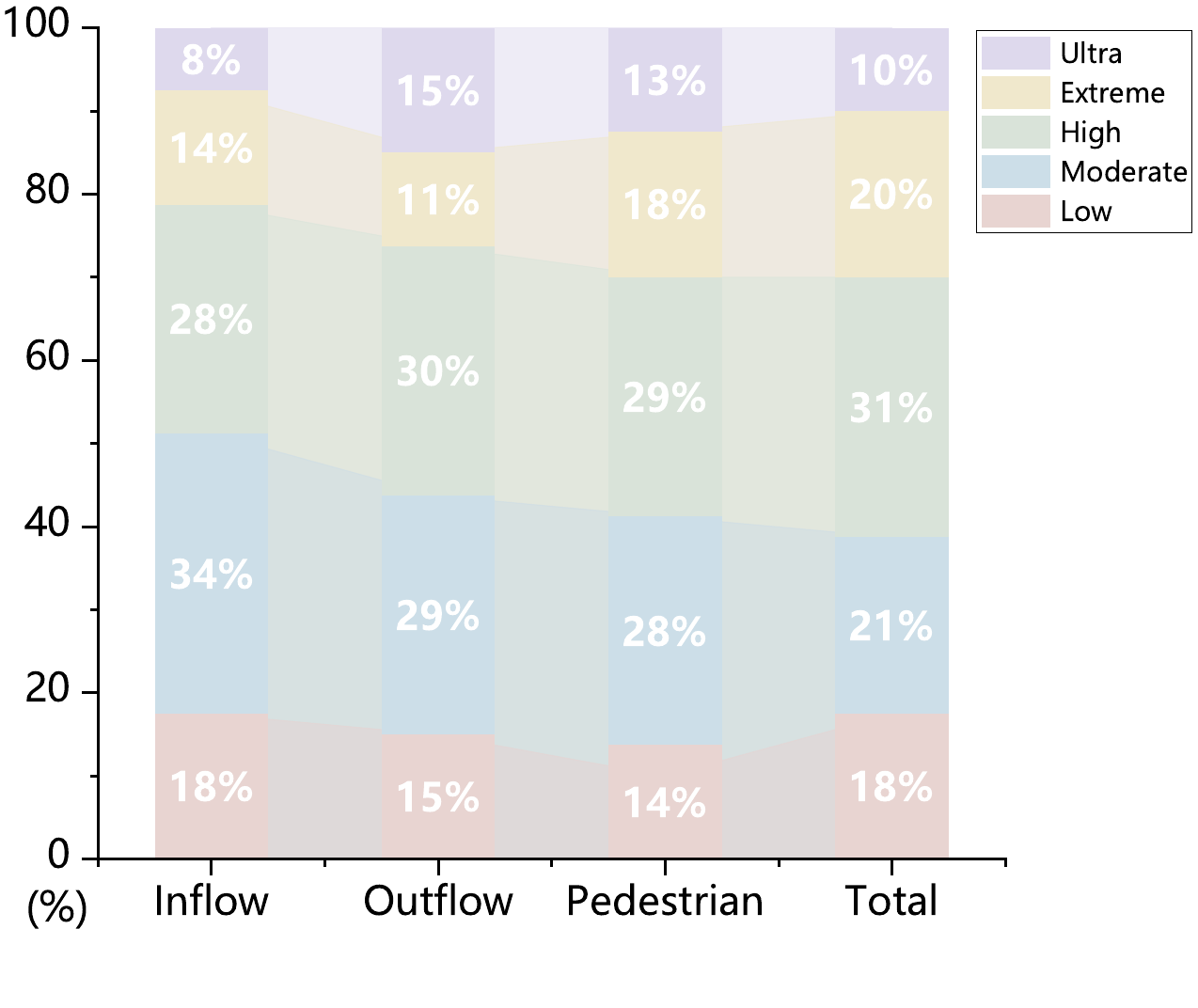}}
	\caption{\textbf{Density and variation distribution of \OurDataset dataset}. (a) depicts the statistical overview of our datasets. Each point denotes a video clip, X-axis represents frame number it contains, Y-axis counts its total annotation number, its scale represents the average density across the video, and the color denotes its belonging scene. (b) shows the dataset partition w.r.t scenes. (c) split class `inflow', `outflow', `pedestrian', and total counts into five density levels. (d) similarly split them into five variation levels, illustrating that a single scene of \OurDataset dataset may cover both static and dynamic situation.}
\label{fig:dataset distribution}
\end{figure*}

\paragraph{Data Collection} We collect $80$ surveillance videos from $15$ metro stations from the Wuhan metro network, as shown in Fig.~\ref{fig:dataset overview}. 
These stations include transfer stations, train stations, commercial districts, and other busy traffic hubs. $11,925$ frames were extracted with two-second sampling intervals from the $80$ videos. 
The longest video sequence is of $864$ frames. 
Video data were collected from $2023$ to $2025$, with a particular focus on densely crowded scenes including peak hours, holidays, and festivals. 
The video resolutions of our dataset are of $720\times576$, $1280\times720$, and $1920\times1080$. 
Fig.~\ref{fig:data_vis} exhibits $7$ representative scenarios encompassed in our dataset: platform, transfer, fare gate, escalator, security, exit/entrance, and lobby, covering both sparse and dense scenes.

\paragraph{Data Annotations} We allocate $10$ well-trained annotators and $2$ checkers to conduct data annotation. X-AnyLabeling\footnote{https://github.com/CVHub520/X-AnyLabeling} is used to annotate each pedestrian head and its inflow and outflow label. 
As shown in Fig.~\ref{fig:data_vis}, by comparing three adjacent frames when labeling the intermediate (current) frame, $223,662$ pedestrians were annotated with four types of labels: 
\romannumeral1) pedestrians that co-exist in the current and next frame are labeled as \textit{pedestrian}; 
\romannumeral2) pedestrians that disappear in the next frame are labeled as \textit{outflow}; 
\romannumeral3) pedestrians that do not appear in the previous frame are labeled as \textit{inflow}; 
\romannumeral4) and pedestrians that are absent in both the previous and next frames, \textit{i.e.}, indicating simultaneous inflows and outflows, are labeled as \textit{both}. 
Additionally, we also provide masks for severely occluded and excessively blurred regions where identities of pedestrians are completely indiscernible, excluding these regions from performance evaluation. 
The density distribution of \OurDataset dataset is shown in Fig.~\ref{fig:dataset distribution}(a), which indicates the relation among annotation number, frame number, crowd density, and scenarios.

\paragraph{Data Partition} To ensure data diversity and mitigate potential biases, we split the dataset (Fig.~\ref{fig:dataset distribution}(b)) into $45$, $15$ and $20$ video sequences separately for training ($60$\%), validation ($20$\%), and testing ($20$\%) with the following criteria in mind: \romannumeral1) different subsets are disjoint, covering different 
video sequences to avoid exactly the same scene between subsets; \romannumeral2) each subset covers 
different scene types and video length; and \romannumeral3) each subset covers 
sufficiently large density variations and flow variations.

\subsection{Benchmark Properties and Statistics}

\paragraph{Extended Long Video Clips} While VIC 
targets cumulative pedestrian inflows within a period of time, the inflow errors also 
accumulate with time. However, existing datasets, particularly those repurposed from MOT benchmarks~\cite{SenseCrowd, HeadHunter-T}, 
mostly host short video clips of tens of seconds (Table~\ref{tab:dataset comparison}). 
Such temporal truncation creates two 
limitations: \romannumeral1) accumulated errors are small 
and fail to reflect 
the long-term counting capability of the model, and \romannumeral2) models trained on 
few-frame sequences can 
have biased representations and 
may perform poorly in scenarios requiring sustained temporal reasoning. 
To overcome these limitations, most videos in our \OurDataset dataset are clipped into $60\sim240$ frames (Fig.~\ref{fig:dataset distribution}(a)), i.e., $2\sim8$ minutes. 

\paragraph{Significant Density and Flow Variations} We characterize density and flow variations with five levels, respectively. Fig.~\ref{fig:dataset distribution}(c) shows five density levels of the three annotated categories and total counts. Considering the scales of pedestrians due to camera angle and imaging height, for `inflow' and `outflow', we 
split the average density (average counts across frames) per sequence into $\tt{sparse}\in \left( 0,1\right]$, $\tt{normal}\in \left( 1,2\right]$, $\tt{crowded}\in \left( 2,3\right]$, $\tt{packed}\in \left( 3,4\right]$, and $\tt{jam}\in \left( 4,+\infty\right]$ with an interval of $1$, and set the interval to $5$ for `pedestrian' and $10$ for total counts. Fig.~\ref{fig:data_vis} also provides some cases of these $5$ different density levels. Notably, the densest sequence on average contains $21$ inflows, $12$ outflows, $230$ pedestrians, and $242$ total counts. 
Considering that, some relatively static frames may 
decrease the average density, we depict the flow variation range of $3$ classes and total counts in each clip to capture the temporal flow variation in each scenario, as shown in Fig.~\ref{fig:dataset distribution}(d). The interval is $4$ for `inflow' and `outflow', $10$ for `pedestrian', and $12$ for total counts. It can be observed that our dataset provides exhaustive coverage of flow variations, ranging from sparse to extremely crowded scenarios. Particularly, the maximum flow variation range is $28$ for `inflow', $28$ for `outflow', $134$ for `pedestrian', and $144$ for total counts. The relative flow variation on average (relative to total pedestrian counts) shown in Table~\ref{tab:dataset comparison} also highlights the dynamic nature of our dataset, which demonstrates that both sparse and dense crowds can be temporally included in one single clip of our dataset. It can be observed that our dataset provides exhaustive coverage of flow variations, ranging from sparse to extremely crowded scenarios.

\paragraph{Severe Occlusions and Scale Variations} Since VIC 
still falls in the generic domain of crowd counting, 
VIC approaches should be 
robust to occlusions and 
scale variations commonly presented in 
crowd counting. 
However, existing VIC data~\cite{SenseCrowd, HeadHunter-T} 
are mostly captured from 
nearly bird-eye 
views; 
the overlap between individuals is minimal and target scales remain consistent, making even the most crowded scenes insufficiently challenging for VIC. Considering the characteristics of surveillance cameras, which typically feature lower heights and smaller viewing angles compared to drone perspectives, occlusion and scale variations are particularly more prevalent in our dataset than in others. Fig.~\ref{fig:data_vis} provides some annotated examples of dense scenarios, where people are severely occluded and their scales vary. 


\section{Proposed Method}
\begin{figure*}[!t]
  \centering
  \centerline{\includegraphics[width=18.2cm]{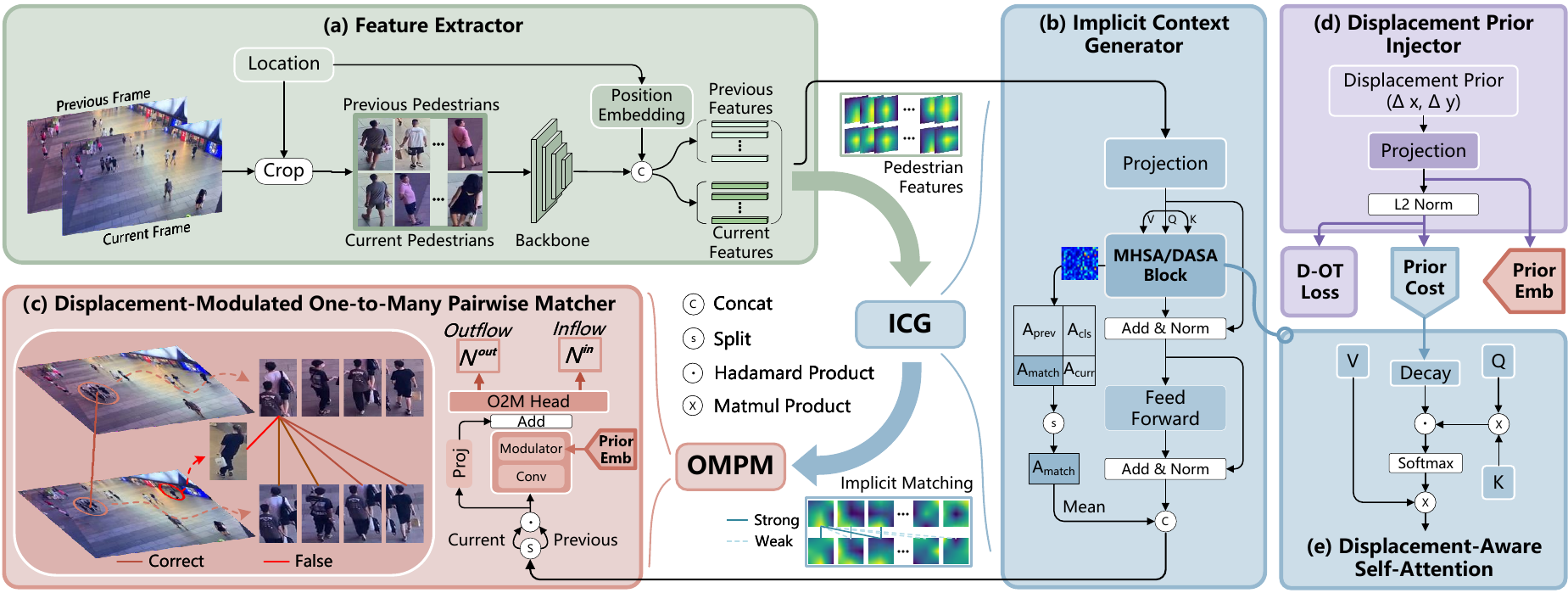}}
  \vspace{-5pt}
  \caption{\textbf{Technical pipeline of OMAN++}. (a) Individuals are first located and cropped with pre-provided location, followed by position-embedded feature extraction. We then apply (b) Implicit Context Generator (ICG) to execute implicit context enhancement with standard Multi-Head Self-Attention (MHSA) or (e) Displacement-Aware Self-Attention (DASA), and resort to (c) One-to-Many Pairwise Matching (OMPM) to implement O2M assignment via a matching head, generating individual-to-group matches and therefore the inflow and outflow numbers. To capture motion patterns of pedestrians, we further devise (d) Displacement Prior Injector (DPI) to inject displacement prior into each module through (e) DASA, a displacement modulator, and a displacement-informed OT (D-OT) loss, keeping final inflow and outflow predictions under the physical constraint.}
\label{fig:Pipeline}
\end{figure*}

\subsection{Problem Formulation}
\label{ssec:prop0}
Given a video sequence $I=\left \{ I_{0}, I_{1}, ..., I_{T} \right \}$ with $T+1$ frames, where the $t$-th frame $I_{t}$ contains $N_{t}$ pedestrians, the video is sampled with an interval of $\sigma$ frames for efficiency. By coupling adjacent frames as pairs $\left \{I_{(k-1)\sigma}, I_{k\sigma} \right \}_{k=1}^{T/\sigma}$, in the $k$-th frame pair, \textit{pedestrians} who do not exist in $ I_{(k-1)\sigma}$, but show up in $I_{k\sigma}$ for the first time, are denoted as \textit{inflows}, with the inflow number $ N_{k\sigma}^{\tt in}$. In contrast, \textit{pedestrians} visible in $I_{(k-1)\sigma}$ but disappeared in $ I_{k\sigma}$ are regarded as \textit{outflows}, with the outflow number $N_{(k-1)\sigma}^{\tt out}$. 
The goal of VIC is to count unique pedestrians from a video sequence, which can be formulated by
\begin{equation}
    \label{eq:VIC}
    N_{\tt total} = N_{0} + \sum_{k=1}^{\tt T/\sigma}  N_{k\sigma}^{\tt in},
\end{equation}
where $N_{\tt total}$ is the total number of unique pedestrians, and $N_{0}$ is the first-frame pedestrian count. 

For ease of exposition, we use $t$ and $t-1$ to indicate the timestamp with an interval of $\sigma$ in what follows. 

\subsection{Overview}
\label{ssec:prop1}
The technical pipeline of OMAN++ is shown in Fig.~\ref{fig:Pipeline}. Following CGNet~\cite{CGNet}, given pedestrian locations inferred by a locator, we first crop pedestrian patches of size $h\times w$ from video frames and 
compute cross-frame location displacements between pedestrians. 
The pedestrian features are then extracted with position embedding, 
followed by an Implicit Context Generator (ICG) used to implicitly provide the group context to inform potential pedestrian matches via multi-head decoupled self-attention. A One-to-Many Pairwise Matcher (OMPM) is used to identify individual-to-group correspondences. 
To exploit the displacement prior, a Displacement Prior Injector (DPI) 
can work through 
a Displacement-Aware Self-Attention (DASA) block (in place of the multi-head self-attention block), a displacement modulator, and an improved Optimal Transport (OT) loss.
In what follows, we first define the problem setting and then explain the technical details.

\subsection{Implicit Context Generator}
\label{ssec:prop2}
Inspired by a recent observation~\cite{CACViT
} that self-attention can be decoupled to capture the implicit token relations for similarity matching, we introduce an ICG module to generate the group context and build the pedestrian relations between frames to inform potential pedestrian matches. 
Formally, let $m=N_{t-1}$ and $n=N_{t}$ denote the pedestrian count in the $(t-1)$-th and the $t$-th frames, respectively. As shown in Fig.~\ref{fig:Pipeline}(b), the encoded pedestrian descriptors are first reshaped and projected into a $d$-dimensional feature vector, denoted by $\bm F_{t}\in\mathbb{R}^{n \times d}$. 
By concatenating $\bm F_{t-1}$ and $\bm F_{t}$, we arrive at $\bm F_{t-1,t}\in\mathbb{R}^{(n+m) \times d} $, which is then fed to the standard multi-head self-attention blocks. 
After self-attention, an attention map $\bm A \in\mathbb{R}^{(n+m) \times (n+m)}$, which indicates the affinities between tokens, can be extracted and split into four sub-attention maps~\cite{CACViT}, namely $\bm A_{\tt prev}\in\mathbb{R}^{m \times m}$,  $\bm A_{\tt cls}\in\mathbb{R}^{m \times n}$,  $\bm A_{\tt match}\in\mathbb{R}^{n \times m}$, and $\bm A_{\tt curr}\in\mathbb{R}^{n \times n}$, taking the form
\begin{equation}
    \label{eq:attention map}
    {\bm A=}\begin{bmatrix}
      \bm A_{\tt prev} & \bm A_{\tt cls}\\
      \bm A_{\tt match} & \bm A_{\tt curr}
    \end{bmatrix}_{(n+m)\times (n+m)}\,,
\end{equation}
where $\bm A_{\tt prev}$ and $\bm A_{\tt curr}$ indicate the pedestrian similarity of the previous frame ($\bm F_{t-1}$) and of the current frame ($\bm F_{t}$), respectively, while $\bm A_{\tt match}$ and $\bm A_{\tt cls}$ both represent the cross-frame correlations among pedestrians. 
To characterize the group context, we average $\bm A_{\tt match}$ and $A_{\tt cls}$ by column to derive $\bar{\bm A}_{\tt match}\in\mathbb{R}^{n}$ and $\bar{\bm A}_{\tt cls}\in\mathbb{R}^{m}$. 
By concatenating both $\bar{\bm A}_{\tt match}$ and $\bar{\bm A}_{\tt cls}$ with the output of self-attention, we acquire the $d^{\prime}$-dimensional context-informed representation $\bm F_{t-1,t}^{^{\prime}}\in\mathbb{R}^{(n+m) \times d^{\prime}}$ indicating the degree of cross-frame pedestrian connections, that is,
\begin{equation}
    \label{eq:ICG}
        {\bm F_{t-1,t}^{^{\prime}}} =  {\tt g}\left({\tt attn}(\bm F_{t-1,t})\oplus (\bar{\bm A}_{\tt match}\oplus \bar{\bm A}_{\tt cls})\right)\,,
\end{equation}
where $\oplus$ is the concatenation operator, ${\tt attn}(\cdot )$ is the multi-head self-attention block, and ${\tt g}(\cdot )$ is a feed-forward layer with residual connections.

\subsection{One-to-Many Pairwise Matcher}
\label{ssec:prop3}
Here we 
introduce the OMPM implementing the O2M assignment. By assigning an individual to a group, one can harness the social grouping prior of pedestrians, especially in crowded scenes. This benefit is also mentioned in some related tasks such as ReID~\cite{Group-reid} and MOT~\cite{TransMOT}. 

We formulate the O2M assignment as a classification problem. 
Concretely, the context-enhanced cross-frame pedestrian representation $\bm F_{t-1,t}^{^{\prime}}$ is first split into single-frame representations  
$\bm F_{t-1}^{^{\prime}}=\{\bm f_{t-1}^{1}, \bm f_{t-1}^{2}, \dots, \bm f_{t-1}^{m}\}\in\mathbb{R}^{m \times d^{\prime}}$ and $\bm F_{t}^{^{\prime}}=\{\bm f_{t}^{1}, \bm f_{t}^{2}, \dots, \bm f_{t}^{n}\}\in\mathbb{R}^{n \times d^{\prime}}$, where $\bm f_{t-1}^{i}\in\mathbb{R}^{d^{\prime}}$ corresponds to the $i$-th pedestrian in the $(t-1)$-th frame, $i=1,...,m$, and $\bm f_{t}^{j}\in\mathbb{R}^{d^{\prime}}$ the $j$-th pedestrian in the $t$-th frame, $j=1,...,n$. 
Then the similarity vector $\bm s_{t-1,t}^{ij}\in\mathbb{R}^{d^{\prime}}$ is computed as $\bm s_{t-1,t}^{ij}=\bm f_{t-1}^{i}\odot \bm f_{t}^{j}$, where $\odot$ is the Hadamard product. 
Finally, $\bm s_{t-1,t}^{ij}$ is fed into an MLP to generate the probability of a candidate match $p_{t-1,t}^{ij}$ using a $\tt sigmoid$ function by
\begin{equation}
    \label{eq:match1}
    p_{t-1,t}^{ij} = {\tt sigmoid}\left({\tt MLP}({\bm s_{t-1,t}^{ij}})\right)\,.
\end{equation}
In this way, the inflow count $N_t^{\tt in}$ of the $t$-th frame and the outflow count $N_{t-1}^{\tt out}$ of the $(t-1)$-th frame can be inferred by
\begin{align}
    \label{eq:match2}
    N_t^{\tt in}&=N_{t}-\sum_{ij}\lfloor p_{t-1,t}^{ij} \rceil\,,\\
    N_{t-1}^{\tt out}&=N_{t-1}-\sum_{ij}\lfloor p_{t-1,t}^{ij}\rceil\,,
\end{align}
where $\lfloor \cdot \rceil$ is the rounding operator that rounds a number to the nearest integer.

During inference, a pedestrian in the $t$-th frame is allowed to match a different one in the $(t-1)$-th frame, as long as they are treated as a group. Conceptually, the differences between the constraints for O2O and O2M matching are
\begin{equation}
    \label{eq:O2O}
    \begin{aligned}
    {\tt O2O:}~~~& {\bm M}\mathbbm{1}_{m}\leq 1,\quad{\bm M}^T\mathbbm{1}_{n}\leq 1\\
    {\tt O2M:}~~~& {\bm M}\mathbbm{1}_{m}\leq \eta
    \end{aligned}\,,
\end{equation}
where $\bm{M}\in\mathbb{R}^{n\times m}$ is the matching matrix, $\mathbbm{1}_{n}$ and $\mathbbm{1}_{m}$ are $n$-dimensional and $m$-dimensional column vectors of ones, and $\eta$ is the maximum size of a group.

\subsection{Displacement Prior Injector}
\label{ssec:prop4}

In the vanilla design of ICG and OMPM, only appearance features are considered to implement O2M matching. 
However, appearance cues can degrade in crowded scenes. Here we propose to incorporate the displacement prior into the pipeline to facilitate robust matching.

To exploit the prior, as shown in Fig.~\ref{fig:Pipeline}(d), the DPI is introduced to guide O2M matching. 
Specifically, let pedestrian locations in the $(t-1)$-th and the $t$-th frame be denoted by two point sets ${\bm X}_{t-1}\in\mathbb{R}^{m\times 2}$ and ${\bm X}_{t}\in\mathbb{R}^{n\times 2}$, respectively. 
We first compute the pairwise cross-frame inter-pedestrian displacement ${\mathcal D}\in\mathbb{R}^{n\times m\times 2}$ by 
\begin{equation}
    \label{eq:prior}
    {\mathcal D}^{ij} ={\bm X}^{j}_{t} - {\bm X}^{i}_{t-1},~i=0,1,...,m,~j=0,1,...,n\,.
\end{equation}
The we apply a nonlinear projection $\phi$ to $\mathcal D$ to adjust feature dimension and generate the displacement prior embedding $\mathcal P \in\mathbb{R}^{n\times m\times \frac{d}{hw}}$, denoted by $\mathcal P = \phi(\mathcal D)$.

In what follows, we 
show through three technical improvements on how to use the prior embedding $\mathcal P$ to improve representation, matching, and learning, respectively.

\begin{figure}[!t]
	\centering  
	\subfigbottomskip=3pt 
	\subfigcapskip=-5pt 
	\subfigure[Prior Cost w.r.t X Displacement]{
		\hspace*{-0.4em}\includegraphics[width=0.48\linewidth]{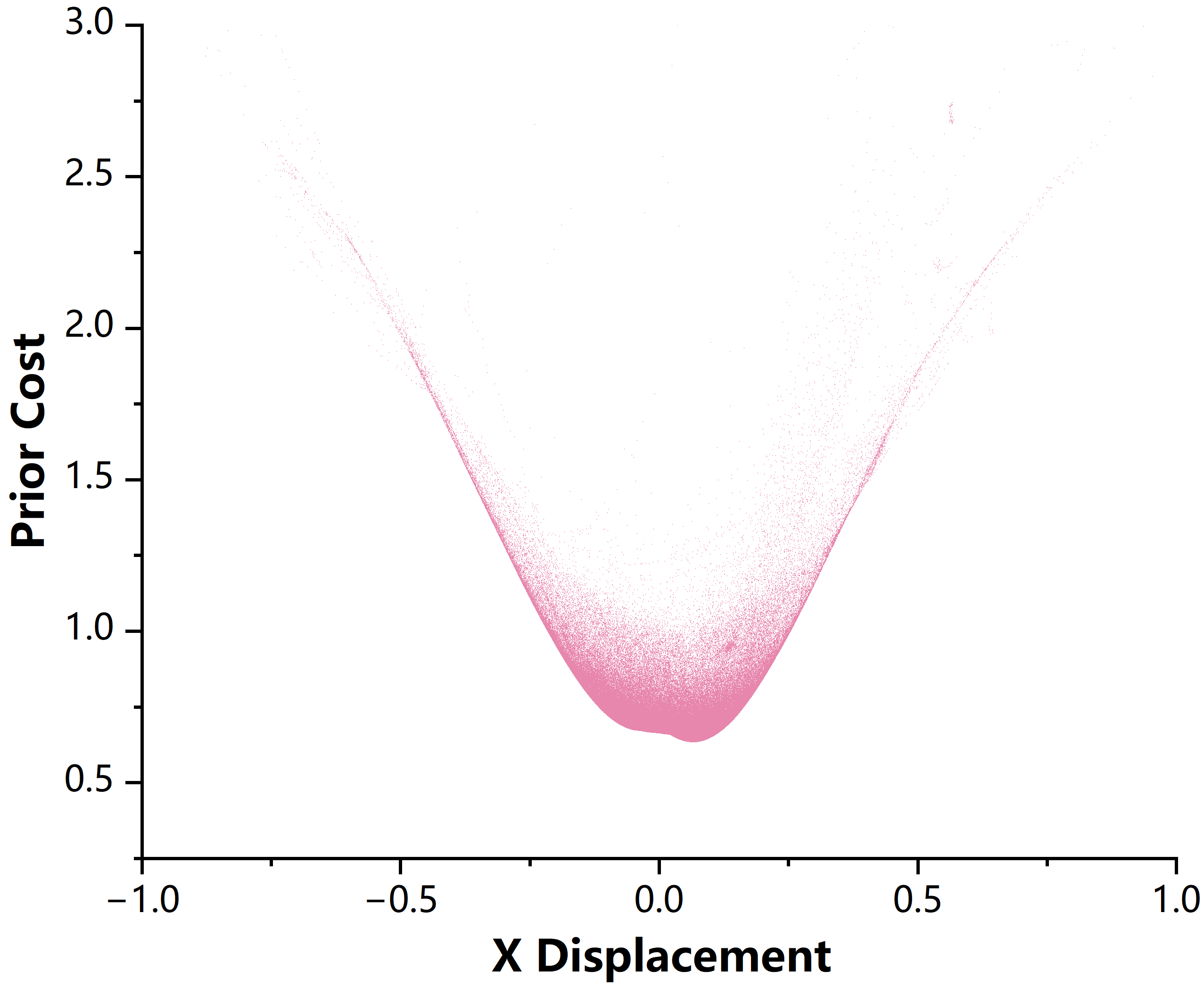}}
	\subfigure[Prior Cost w.r.t Y Displacement]{
		\hspace*{-0.4em}\includegraphics[width=0.48\linewidth]{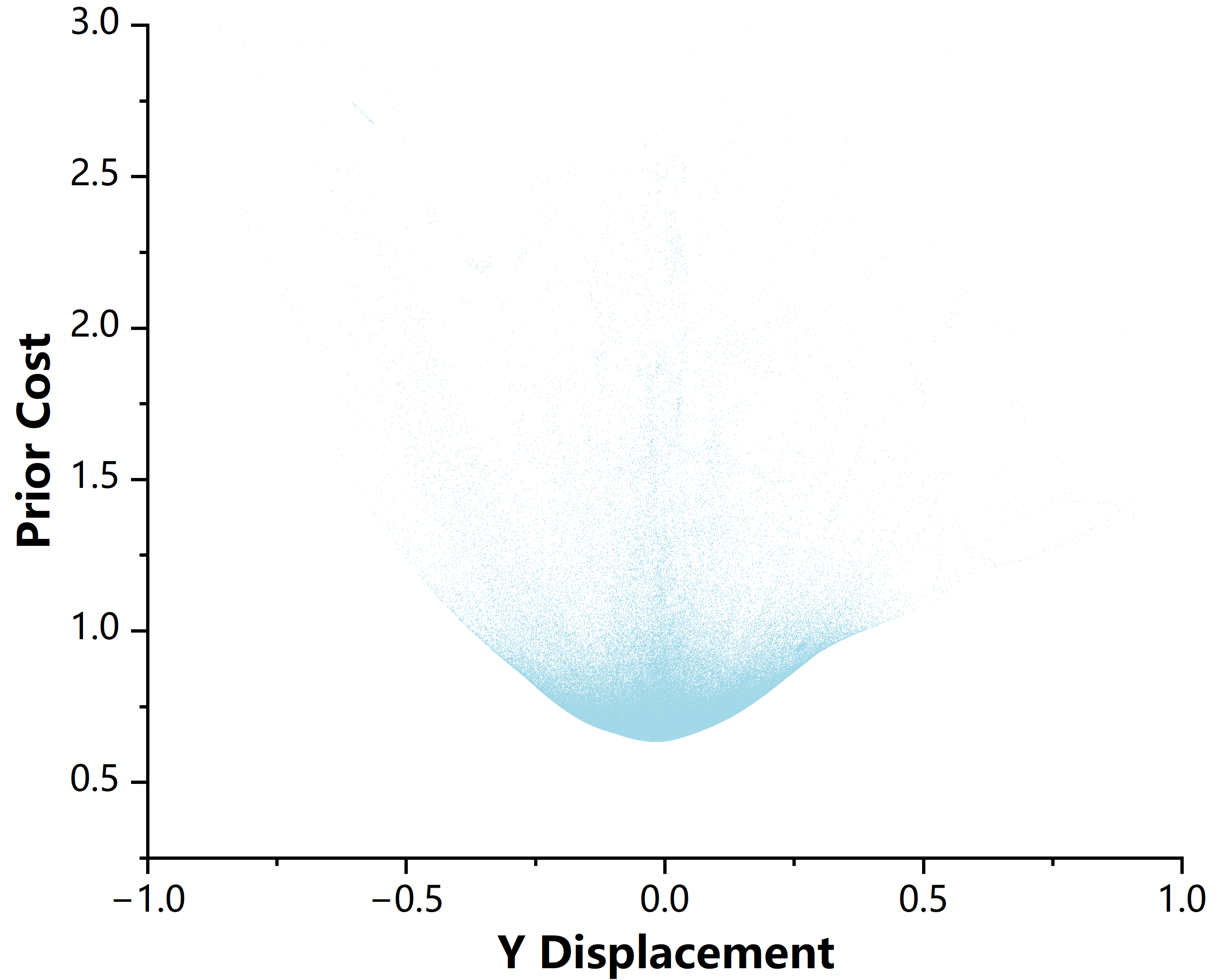}}
	\caption{\textbf{Relationship between the prior cost and the displacement.} Each point denotes a matching pair. The pedestrian displacements are separately shown along the (a) $x$-axis and (b) $y$-axis. The prior cost is generally positively correlated to the absolute value of displacement in both $x$ and $y$ direction, while the trends differ that the prior cost increase faster when pedestrians in frames move horizontally than move vertically.}
\label{fig:cost_wrt_disp}
\end{figure}

\paragraph{Displacement-Aware Self-Attention}
\begin{figure}[!t]
	\centering  
	\subfigbottomskip=3pt 
	\subfigcapskip=-5pt 
	\subfigure[Visualization of ICG w/o Displacement Prior.]{
		\hspace*{-0.4em}\includegraphics[width=1\linewidth]{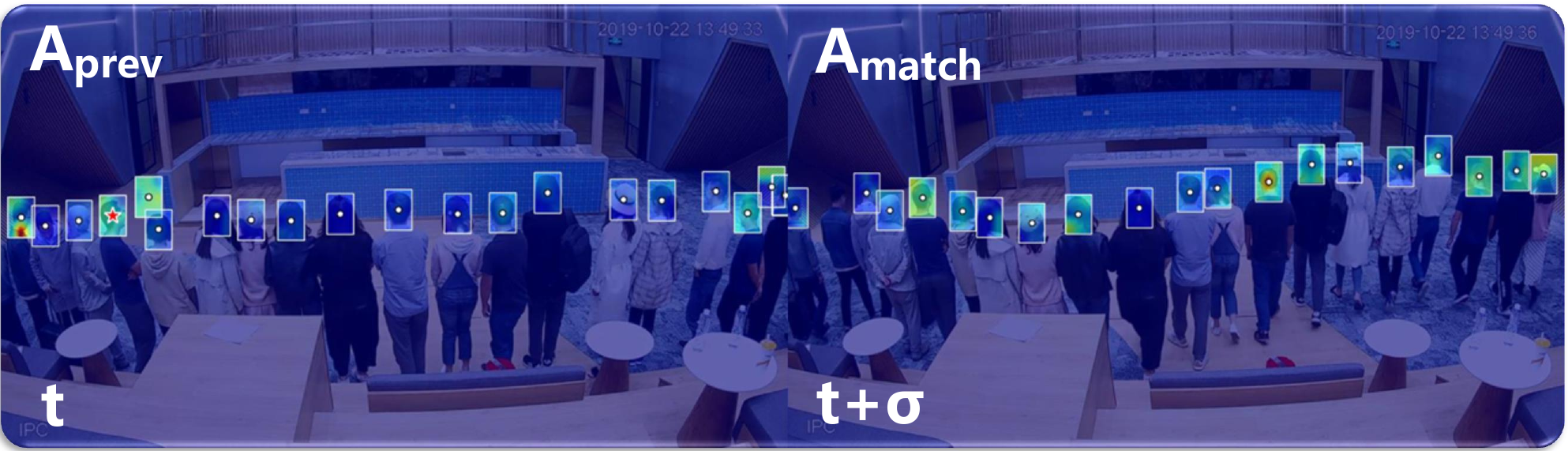}}
	\subfigure[Visualization of ICG w/ Displacement Prior.]{
		\hspace*{-0.4em}\includegraphics[width=1\linewidth]{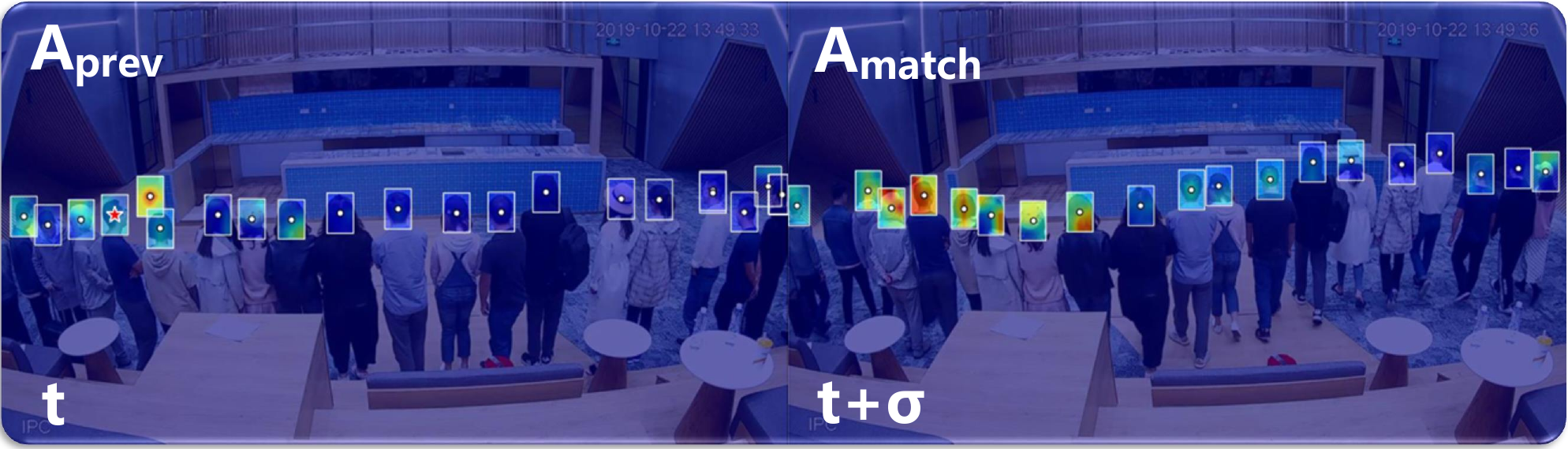}}
	\caption{\textbf{Visualization of displacement guided attention.} (a) shows the power of vanilla self-attention, and (b) depicts how we apply displacement prior to guide self-attention, where white boxes are crop patches with attention map inside. The pedestrian with red star in $t$-th frame serves as a candidate to search in the $(t+\sigma)$-th frame.}
\label{fig:displacement visualization}
\end{figure}

In Section~\ref{ssec:prop2}, we simply leverage a standard MHSA block. 
Inspired by \cite{DFormerv2}, here we introduce a novel displacement-aware self-attention (DASA) block to incorporate motion cues.
In the ICG module, the pedestrian representation $\bm F_{t-1,t}$ is normalized and projected to the query $\bm Q \in\mathbb{R}^{(n+m)\times d}$, key $\bm K \in\mathbb{R}^{(n+m)\times d}$ and value $\bm V \in\mathbb{R}^{(n+m)\times d}$, and are then fed into DASA in Fig.~\ref{fig:Pipeline}(e). 
The prior embedding $\mathcal P$ is then used to compute the prior cost $\bm \gamma$, modulating the query-key product via the Hadamard product, which takes the form
\begin{equation}
    \label{eq:ICG-d1}
    \begin{aligned}
        &\bm Q = \bm W_{Q}\bm F_{t-1,t}\,,~~~~ \bm K = \bm W_{K}\bm F_{t-1,t}\,,\\
        &\bm V = \bm W_{V}\bm F_{t-1,t}\,,~~~~ {\bm \gamma} = \Vert \mathcal P \Vert_2\,,\\
        &{\tt DASA}(\bm F_{t-1,t}) = \left({\tt Softmax}(\frac{\bm Q\cdot \bm K^\top}{\sqrt{d} } )\odot \theta^{\bm \gamma}\right)\bm V\,,\\
    \end{aligned}
\end{equation}
where $\bm W_Q$, $\bm W_K$ and $\bm W_V$ are learnable parameters, $\Vert \cdot \Vert_2$ is the $\ell_2$ norm across all channels, ${\tt Softmax}(\cdot)$ is the softmax function, 
and $\theta$ is a learnable parameter. To confirm the difference between the prior cost and the displacement itself in both directions and amplitudes, we explore their relationship in Fig.~\ref{fig:cost_wrt_disp}, showing two clear but different quadratic dependences of displacements along the two axes. 

We also compare the attention map between MHSA and DASA in Fig.~\ref{fig:displacement visualization} to highlight the guidance of the displacement prior. The ICG module fetches $\bm F_{t-1,t}^{^{\prime}}$ and provides implicit context for OMPM.
With DASA, Eq.~\eqref{eq:ICG} amounts to
\begin{equation}
    \label{eq:ICG-d2}
        {\bm F_{t-1,t}^{^{\prime}}} = {\tt g}\left({\tt DASA}(\bm F_{t-1,t})\oplus (\bar{\bm A}_{\tt match}\oplus \bar{\bm A}_{\tt cls})\right)\,,
\end{equation}
where ${\tt DASA}(\cdot )$ is our proposed DASA block, and ${\tt g}(\cdot )$ is a feed-forward layer with residual connections.

\paragraph{Displacement-Modulated Matcher}
For dense crowds, pedestrian motion is usually suppressed. Thus, the background and group context should be emphasized for matching stability, and vice versa for sparse crowds. 
Consequently, we introduce a coarse-to-fine modulator in Fig.~\ref{fig:Pipeline}(c) to adaptively weight the contributions of appearance features and motion-based geometric cues, superseding the standard MLP in Section~\ref{ssec:prop3}.

Given the similarity representation $\bm s_{t-1,t}^{ij}$, a linear projection $\delta(\cdot)$ and layer normalization ${\tt LN}(\cdot)$ are first applied to form a representation $\bm z_{\tt coarse}\in\mathbb{R}^{\frac{d}{hw}}$ specific to coarse modulation, which captures global similarity but lacks fine-grained spatial details with explicit structural modeling, defined by
\begin{equation}
    \label{eq:coarse}
    {\bm z_{\tt coarse}}={\tt LN}\left(\delta(\bm s_{t-1,t}^{ij})\right)\,.
\end{equation}
In addition, we reshape $\bm s_{t-1,t}^{ij}$ to a 3D tensor $\mathcal S_{t-1,t}^{ij} \in\mathbb{R}^{h\times w\times \frac{d^{\prime}}{hw} }$ to restore the local spatial context. 
To incorporate the displacement prior, we broadcast the prior embedding $\mathcal P$ into $\mathcal P^{\prime}\in \mathbb{R}^{h\times w\times \frac{d}{hw}}$, and use $\mathcal P^{\prime}$ to modulate the $(\frac{d}{hw})$-channel convolutional response of $\mathcal S_{t-1,t}^{ij}$ and generate another representation $\bm z_{\tt fine}\in\mathbb{R}^{h\times w\times \frac{d}{hw}}$ w.r.t.\ fine modulation, allowing the similarity map to adapt its response conditioned on the displacement. This procedure can be formulated by 
\begin{equation}
    \label{eq:fine}
    \begin{aligned}
    &{\bm z_{\tt fine}}=\alpha(\mathcal P^{\prime})\odot {\tt ReLU}\left({\tt Conv}(\mathcal S_{t-1,t}^{ij})\right)+\beta(\mathcal P^{\prime})\,,\\
    \end{aligned}
\end{equation}
where ${\tt Conv}(\cdot)$ denotes a convolution layer, $\alpha(\cdot)$ and $\beta(\cdot)$ are linear layers, and $\odot$ is the Hadamard product, ${\tt ReLU}$ is the rectified linear unit activation function. Then we obtain the fused representation ${\bm z_{\tt fused}}$ by
\begin{equation}
    \label{eq:coarse2fine}
    {\bm z_{\tt fused}}={\bm z_{\tt coarse}}+{\tt GAP}({\bm z_{\tt fine}})\,,
\end{equation}
where ${\tt GAP}(\cdot)$ denotes global average pooling. With the guidance of displacements at both feature and matching levels, the fused representation $\bm z_{\tt fused}$ can then be fed into an O2M Head to generate the final inflow and outflow counts. That is, Eq.~\eqref{eq:match1} can be updated to
\begin{equation}
    \label{eq:match_fuse}
    p_{t-1,t}^{ij} = {\tt sigmoid}\left({\tt MLP}({\bm z_{\tt fused}})\right)\,.
\end{equation}

\paragraph{Displacement-Informed Optimal Transport Loss}
Following CGNet~\cite{CGNet}, we employ the Optimal Transport (OT) loss $\mathcal{L}_{\tt OT}$ for group-level contrastive learning. For appearance learning, we take the same cost matrix $\bm C_{\tt appear} \in\mathbb{R}^{m\times n}$ as CGNet, which is computed by similarity matrices. For displacement learning, we use the prior cost ${\bm \gamma\in\mathbb{R}^{m\times n}}$ generated by ICG (explained in \ref{ssec:exp2}) to generate the displacement cost matrix $\bm C_{\tt disp}\in\mathbb{R}^{m\times n}$, defined by
\begin{equation}
    \label{eq:disp cost}
        \bm C_{\tt disp} = 1-\frac{\exp(\hat{{\bm \gamma}}_{ij})}{\sum_k\exp(\hat{{\bm {\bm \gamma}}}_{kj})+\sum_k\exp(\hat{{\bm \gamma}}_{ik})-\exp(\hat{{\bm \gamma}}_{ij})}\,,
\end{equation}
where $\hat{{\bm \gamma}}_{ij}=\frac{{\bm \gamma}_{ij}-\mu_{{\bm \gamma}}}{\sigma_{{\bm \gamma}}+\epsilon}$, $\mu_{{\bm \gamma}}$ and $\sigma_{{\bm \gamma}}$ are the mean and standard deviation of the prior cost ${\bm \gamma}$, $\epsilon$ is a smoothing term to avoid zero division, and $i$, $j$ and $k$ are indices. Finally, we arrive at the displacement-informed OT loss
\begin{equation}
    \label{eq:OT}
    \begin{aligned}
        &\mathcal{L}_{\tt D-OT}=-\min_{\bm \Pi}\sum_{i=1}^n\sum_{j=1}^m{\bm C}_{ij}{\bm \Pi}_{ij}\,,\\
        &\bm C=\lambda \bm C_{\tt disp} + (1-\lambda) \bm C_{\tt appear}\,,
    \end{aligned}
\end{equation}
where $\bm \Pi$ is the matching matrix computed by Sinkhorn algorithm~\cite{sinkhorn}, and $\lambda$ is a hyper-parameter. Notably, $C_{\tt disp}$ is not only designed for displacement prior but also enhances the discriminability of appearance features by pushing 
away representations of distant pedestrians and by pulling 
close those of proximity during OT solving process.

\subsection{Loss Function}
\label{ssec:prop5}
Besides displacement-informed OT loss, the cross entropy loss $\mathcal{L}_{\tt cls}$ is used to supervise the training of OMPM. For training efficiency, we use Hungarian algorithm~\cite{Hungarian} to select candidates that contain both correct matching pairs and incorrect pairs, and pairwise matching pairs are then fed to a modulator and an MLP. Here, for a pedestrian in a frame, we group this pedestrian and its neighboring ones within a certain normalized distance, $0.2$ for example (according to Fig.~\ref{fig:priors}(b)), in adjacent frames. Then all pairwise matches within this group are considered positive matches, while all remaining ones are negative samples. 
Finally, the total loss $\mathcal{L}_{\tt total}$ used by OMAN++ takes the form
\begin{equation}
    \label{eq:loss}
\mathcal{L}_{\tt total} = \mathcal{L}_{\tt D-OT} + \mathcal{L}_{\tt cls}\,.
\end{equation}

\section{Experiments}
To quantitatively evaluate the effectiveness of OMAN++, we conduct extensive experiments on three standard VIC benchmarks and our \OurDataset dataset.
\subsection{Datasets and Metrics}
\label{ssec:exp1}
\textbf{Datasets.} Following previous work~\cite{DRNet,MDC}, we choose the SenseCrowd~\cite{SenseCrowd}, CroHD~\cite{HeadHunter-T} and MovingDroneCrowd~\cite{MDC} datasets, and we also compare methods that require no ID annotation on our \OurDataset dataset. SenseCrowd consists of $634$ videos, with head coordinates and ID annotations. CroHD is an MOT dataset with $9$ videos ($4$ for training, and $5$ for testing), involving crowded indoor and outdoor scenes. MovingDroneCrowd is a recent VIC dataset with ID annotations of $89$ videos. For VIC, we transform the ID annotations into inflow/outflow labels.

\vspace{5pt}
\textbf{Metrics.} We use the standard VIC metric Weighted Relative Absolute Errors (WRAE) to quantify the performance, defined by
\begin{equation}
    \label{eq:WRAE}
    WRAE=\sum_{i=1}^{K}\frac{T_{i}}{\sum_{j=1}^{K}T_{j}}\frac{|N_{i}-\hat{N}_{i}|}{N_{i}}\times100\%\,,
\end{equation}
where $K$ represents the video number, $T_{i}$ denotes the length of the $i$-th video sequence, $N_{i}$ is the ground-truth pedestrian count of the $i$-th video, and $\hat{N}_{i}$ is the predicted one.
Since VIC also closely relates to counting, generic counting metrics such as Mean Absolute Error (MAE) and Mean Squared Error (MSE) are reported as well.

\subsection{Implementation Details }
\label{ssec:exp2}

\textbf{Training Details.} For pedestrian localization, we adopt the framework of Point quEry Transformer (PET)~\cite{PET}. For a fair comparison with CGNet~\cite{CGNet}, the ConvNext-S~\cite{convnext} is used to encode pedestrians into descriptors. 
The learning rate and weight decay are both set to $1e^{-4}$, $\lambda$ is set as $0.1$ and all other training details are kept the same as CGNet~\cite{CGNet}. Our model is trained on $4$ RTX $3090$ GPUs with the batch size set to $4$.

\vspace{5pt}
\textbf{Testing Details.} 
Following previous work~\cite{DRNet, MDC}, the frame interval $\sigma$ is set to $75$ for SenseCrowd~\cite{SenseCrowd}, $15$ for CroHD~\cite{HeadHunter-T} dataset, $4$ for MovingDroneCrowd~\cite{MDC} dataset, and $1$ for \OurDataset dataset.

\begin{table}[!t]  \footnotesize
    \centering
    \caption{\textbf{Results on SenseCrowd dataset.} D$0$$\sim$D$4$ represents different crowd densities, ranging [$0$,$50$), [$50$,$100$), [$100$,$150$), [$150$,$200$) and [$200$,$+\infty$). Best performance is in \textbf{boldface}, and second best is \underline{underlined}}
    \label{tab:SENSE}
    \renewcommand{\arraystretch}{1} 
    \addtolength{\tabcolsep}{-4.3pt} 
    \begin{tabular}{@{}lcccccccc@{}}
        \toprule
        \multirow{2}{*}{Method} & \multirow{2}{*}{MAE$\downarrow$} & \multirow{2}{*}{MSE$\downarrow$} & \multirow{2}{*}{WRAE(\%)$\downarrow$} & \multicolumn{5}{c}{Density\textbf{ }level (MAE$\downarrow$)} \\
        \cmidrule{5-9}
        & & & & D$0$ & D$1$ & D$2$ & D$3$ & D$4$ \\
        \midrule
        FairMOT~\cite{FairMOT} & 35.4 & 62.3 & 48.9 & 13.5 & 22.4 & 67.9 & 84.4 & 145.8 \\
        HeadHunter-T~\cite{HeadHunter-T} & 30.0 & 60.6 & 44.9 & 8.9 & 17.3 & 56.2 & 96.4 & 131.4 \\
        \midrule
        LOI~\cite{LOI-2016} & 24.7 & 33.1 & 37.4 & 12.5 & 25.4 & 39.9 & 104.1 & 115.2 \\
        \midrule
        DRNet~\cite{DRNet} & 12.3 & 24.7 & 12.7 & 4.1 & 8.0 & 23.3 & 50.0 & 77.0 \\
        CGNet~\cite{CGNet} & 8.9& 17.7& 12.6 & 5.0 & \textbf{5.8} &\textbf{8.5} & 25.0 & 63.4 \\ 
        FMDC~\cite{FMDC} & 14.3 & 28.9 & 14.9 & 4.8 & 8.8 & 22.5 & 54.1 & 88.8 \\
        PDTR~\cite{PDTR} & 9.6 & 17.6 & 11.4 & 4.6 & 6.8 & 14.7 & 23.6 & 60.6 \\ 
        MDC~\cite{MDC} & 8.6 & 15.9 & \underline{11.0} & 4.0 & 6.6 & 12.6 & 24.9 & \textbf{47.9} \\
        \midrule
        OMAN (Ours) & \underline{8.3} & \underline{15.4} & 11.1 & \underline{3.9} & \underline{6.2} & 13.2 & \underline{20.3} & \underline{48.2}\\
        OMAN++ (Ours) & \textbf{7.7} & \textbf{14.6} & \textbf{10.4} & \textbf{3.3} & 6.3 & \underline{12.3} & \textbf{12.5} & \textbf{47.9} \\
        \bottomrule
    \end{tabular}
\end{table}

\begin{table}[!t]  \footnotesize
    \centering
    \caption{\textbf{Results on MovingDroneCrowd dataset.} D$0$$\sim$D$3$ represents different crowd densities, ranging [$0$,$150$), [$150$,$300$), [$300$,$450$) and [$450$,$+\infty$). Best performance is in \textbf{boldface}, and second best is \underline{underlined}}
    \label{tab:MovingDroneCrowd}
    \renewcommand{\arraystretch}{1} 
    \addtolength{\tabcolsep}{-3.1pt} 
    \begin{tabular}{@{}lccccccc@{}}
        \toprule
        \multirow{2}{*}{Method} & \multirow{2}{*}{MAE$\downarrow$} & \multirow{2}{*}{MSE$\downarrow$} & \multirow{2}{*}{WRAE(\%)$\downarrow$} & \multicolumn{4}{c}{Density\textbf{ }level (MAE$\downarrow$)} \\
        \cmidrule{5-8}
        & & & & D$0$ & D$1$ & D$2$ & D$3$\\
        \midrule
        ByteTrack~\cite{FairMOT} & 153.2 & 227.6 & 63.8 & 83.4 & \underline{24.0} & 325.0 & 441.3\\
        BoT-SORT~\cite{HeadHunter-T} & 150.6 & 223.5 & 62.5 & 82.5 & \textbf{22.0} & 327.0 & 430.0\\
        OC-SORT~\cite{OCsort} & 203.6 & 276.8 & 87.8 & 101.5 & 232.0 & 405.0 & 569.3\\
        \midrule
        LOI~\cite{LOI-2016} & 241.8 & 337.9 & 99.6 & 110.1 & 294.5 & 467.6 & 719.3\\
        \midrule
        DRNet~\cite{DRNet} & 81.1 & 126.3 & 33.4 & 28.7 & 129.9 & 217.1 & 246.7\\
        CGNet~\cite{CGNet} & 66.1 & 110.4 & 29.2 & 25.9 & 111.0 & 144.0 & 199.0\\
        FMDC~\cite{FMDC} & 120.3 & 183.6 & 48.8 & 61.7 & 75.7 & 54.9 & 411.1\\ 
        MDC~\cite{MDC} & 41.0 &  \underline{58.3} & 19.3 &  \underline{23.7} & 79.8 &  \underline{41.2} &  \underline{102.9}\\
        \midrule
        OMAN (Ours) & \underline{37.9} & 61.5 &  \underline{18.9} & \textbf{16.4} & 31.0 & 52.0 & 129.0\\
        OMAN++ (Ours) & \textbf{36.1} & \textbf{56.6} & \textbf{17.2} & 25.4 & 73.0 & \textbf{15.0} & \textbf{77.3}\\
        \bottomrule
    \end{tabular}
\end{table}

\begin{table*}[!t] \small
    \centering
    \caption{\textbf{Results on CroHD dataset.} CroHD$11$$\sim$CroHD$15$ represents the five scenes, with $133$, $737$ ,$734$, $1040$, and $321$ as ground-truth counts, respectively. Values in `( )' represents the deviation from ground-truth counts, and `--' is due to unreleased code. Best performance is in \textbf{boldface}, and second best is \underline{underlined}}
    \label{tab:CroHD}
    \renewcommand{\arraystretch}{1} 
    \addtolength{\tabcolsep}{1.5 pt}
    \begin{tabular}{@{}lccccccccc@{}}
        \toprule
        \multirow{3}{*}{Method} & \multirow{3}{*}{MAE$\downarrow$} & \multirow{3}{*}{MSE$\downarrow$} & \multirow{3}{*}{WRAE(\%)$\downarrow$} & \multicolumn{5}{c}{MAE$\downarrow$ on the five testing scenes} \\
        \cmidrule{5-9}
        & & & & CroHD$11$ & CroHD$12$ & CroHD$13$ & CroHD$14$ & CroHD$15$ \\
        & & & & 133 & 737 & 734 & 1040 & 321 \\
        \midrule
        FairMOT~\cite{FairMOT} & 256.2 & 300.8 & 34.1 & 144 (11) & 1164 (427) & 1018 (284) & 632 (408) & 472 (1000) \\
        HeadHunter-T~\cite{HeadHunter-T} & 253.2 & 351.7 & 32.7 & 198 (65) & 636 (101) & 219 (515) & 458 (582) & \textbf{324 (3)} \\
        ByteTrack~\cite{ByteTrack} & 213.2 & 340.2 & 30.8 & 160 (27) & \textbf{761 (24)} & 1467 (733) & 897 (143) & 460 (139) \\
        \midrule
        LOI~\cite{LOI-2016} & 305.0 & 371.1 & 46.0 & 72.4 (60) & 493.1 (243) & 275.3 (458) & 409.2 (630) & 189.8 (131) \\
        \midrule
        DRNet~\cite{DRNet} & 141.1 & 192.3 & 27.4 & 164.6 (31) & 1075.5 (338) & 752.8 (18) & 784.5 (255) & 382.3 (61) \\
        CGNet~\cite{CGNet} & 75.0 & 95.1 & 14.5 &-- (7) & -- (72) & \underline{-- (14)} & -- (144) & -- (138) \\
        FMDC~\cite{FMDC} & 54.2 & 61.7 & 10.7 & \underline{138.9 (5)} & 664.3 (73) & 818.0 (84) & \underline{1005.8 (35)} & 394.9 (73) \\ 
        PDTR~\cite{PDTR} & 60.6 & 73.7 & 12.7 & 109 (24) & \underline{678 (59)} & \textbf{729 (5)} & 935 (105) & 431 (110) \\ 
        MDC~\cite{MDC} & 128.9 & 160.0 & 19.7 & 103.0 (30) & 801.9 (64) & 577.2 (157) & 740.8 (299.2) & 414.8 (93)\\
        \midrule
        OMAN (Ours) & \underline{38.6} & \underline{50.5} & \underline{8.3}& 142 (9) & 636 (101) & 715 (19) & 1003 (37) & \underline{348 (27)} \\ 
        OMAN++ (Ours) & \textbf{35.4} & \textbf{48.8} & \textbf{7.9}& \textbf{138 (5)} & 639 (98) & 698 (36) & \textbf{1048 (8)} & 291 (30) \\     
        \bottomrule
    \end{tabular}
\end{table*}

\begin{table*}[!h] \small
    \centering
    \label{tab:Metro}
    \caption{\textbf{Results on WuhanMetroCrowd dataset}. Platform, Transfer, Fare Gate, Lobby, Escalator, and Exit/Entrance represent six typical scenes in test subset of our dataset. Best performance is in \textbf{boldface}, and second best is \underline{underlined}}
    \renewcommand{\arraystretch}{1} 
    \addtolength{\tabcolsep}{2pt} 
    \begin{tabular}{@{}lcccccccccc@{}}
        \toprule
        \multirow{2}{*}{Method} & \multirow{2}{*}{MAE$\downarrow$} & \multirow{2}{*}{MSE$\downarrow$} & \multirow{2}{*}{WRAE(\%)$\downarrow$}& \multicolumn{6}{c}{WRAE(\%)$\downarrow$ on different scenes}\\
        \cmidrule{5-10}
        &&&&Platform&Transfer&Fare Gate&Lobby&Escalator&Exit/Entrance\\
        \midrule
        ByteTrack\cite{ByteTrack} & 215.5 & 477.8 & 48.3 & 43.7 & 53.9 & 21.5 & 90.2 & 77.9 & 55.3 \\
        OC-Track\cite{OCsort} & 287.9 & 549.6 & 66.8 & 46.9 & 76.6 & 56.4 & 94.2 & 89.8 & 71.2 \\
        \midrule
        CGNet\cite{CGNet} & 172.4 & 368.5 & 37.5 & 33.0 & 51.2 & 26.4 & \underline{63.4} & 47.6 & 48.3 \\ 
        MDC~\cite{MDC} & 166.0 & 439.8 & 32.0 & \underline{18.3} & \textbf{20.4} & 31.6 & 71.2 & 75.8 & \textbf{23.1}  \\
        \midrule
        OMAN (Ours) & \underline{135.9} & \underline{284.1} & \underline{31.8} & 31.6 & 49.5 & \underline{15.8} & 64.2 & \underline{34.9} & \underline{37.8} \\
        OMAN++ (Ours) & \textbf{87.1} & \textbf{160.2} & \textbf{19.8} & \textbf{15.4} & \underline{28.9} & \textbf{11.5} & \textbf{43.5} & \textbf{23.3} & 45.4 \\
        \bottomrule
    \end{tabular}
\end{table*}

\subsection{Experimental Results }
\label{ssec:exp3}

\begin{figure}[!t]
    \centering
    \centerline{\includegraphics[width=1\linewidth]{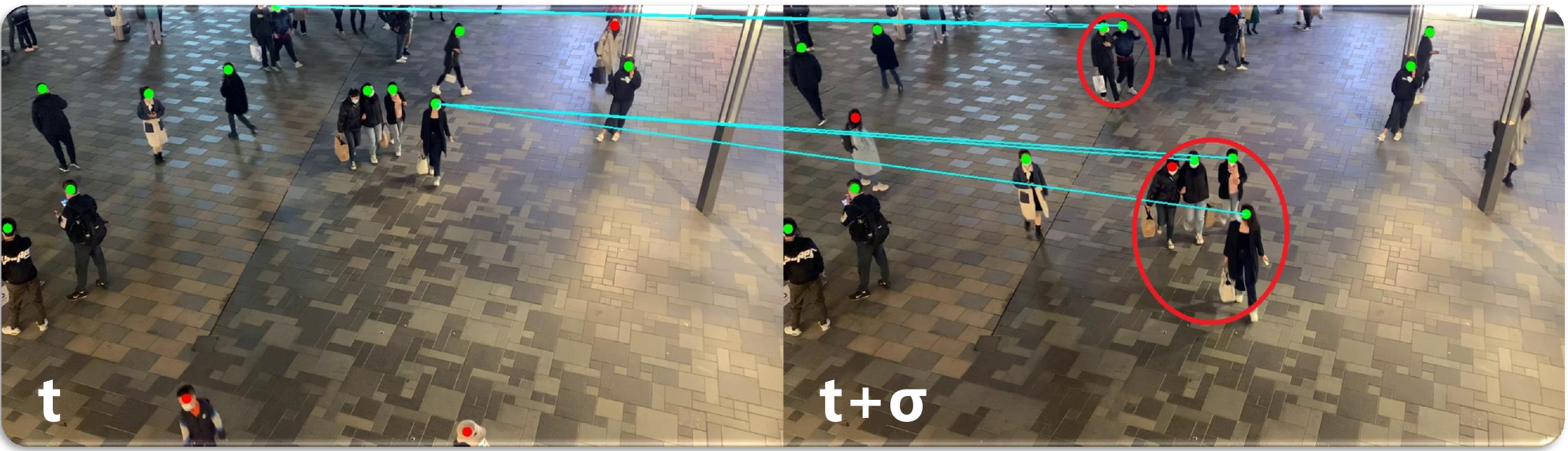}}
    \vspace{-5pt}
    \caption{\textbf{Visualization of O2M matching.} Red dots indicate inflows/outflows, green dots indicate shared pedestrians, red circles represent groups, and blue lines indicate inferred O2M matches.}
    \label{fig:vis_O2M}
\end{figure}


\textbf{Results on SenseCrowd.} Results are reported in Table~\ref{tab:SENSE}. 
MOT and LOI methods almost fail on all density levels, and errors become even larger in denser scenes, while VIC methods significantly reduce the errors. 
The results show that OMAN++ surpasses the VIC baselines in general in all metrics, achieving $7.7$ MAE, $14.6$ MSE, and $10.4$\% WRAE. MAE, MSE and WRAE are $10.5\%$, $8.2\%$, and $5.5\%$ lower than MDC~\cite{MDC}, while $7.2\%$, $5.2\%$, and $5.1\%$ than OMAN~\cite{OMAN}. 
Specifically, our model reports the best results on D$0$, D$3$ and D$4$, including both the sparsest and the densest scenes, and is also with the second-best results achieved on D$2$. 
The results demonstrate that, with a deeper insight of VIC, the one-to-many matching and the motion awareness of OMAN++ benefit matching on different density levels, especially in dense scenes. 
In addition, Fig.~\ref{fig:vis_O2M} visualizes that OMAN++ indeed implements the O2M matching. 

\vspace{5pt}
\textbf{Results on MovingDroneCrowd.} As shown in Table \ref{tab:MovingDroneCrowd}, OMAN++ also outperforms all other VIC methods on the MovingDroneCrowd dataset. OMAN++ achieve the lowest MAE, MSE, WRAE, which are $13.6\%$, $2.9\%$ and $10.9\%$ lower than MDC~\cite{MDC}, and are $4.7\%$, $8.0\%$ and $9.0\%$ lower than OMAN~\cite{OMAN}. Among the four density levels, OMAN++ outperforms on two most crowded scenes, D$2$ and D$3$. The results show a great superiority in dense crowds over existing methods, and also exhibit a significant robustness to sparse scenes and to dynamic challenges caused by moving drones.
However, we also observe that OMAN++ exhibits a noticeable performance degradation on D$1$, compared to OMAN and some MOT methods. The decline may be largely due to the limited test data on D$1$ (only one single video sequence) and the small motion of pedestrians.

\vspace{5pt}
\textbf{Results on CroHD.} Experimental results on the CroHD dataset are shown in Table \ref{tab:CroHD}. We observe that our proposed OMAN++ is also the best performing one. Generally, MAE, MSE and WRAE are respectively reduced by $8.3$\%, $3.4$\%, and $5.1$\% when compared to OMAN~\cite{OMAN}, and reduced by $34.7\%$, $20.9\%$, and $26.2\%$ when compared to FMDC~\cite{FMDC}. 
CroHD$11$$\sim$CroHD$15$ represent five different scenes with different densities, while the differences between predicted and true values in different scenes are generally small, with the lowest MAE on CroHD$11$ and CroHD$14$, indicating that our model is robust to various densities of the pedestrian.

\begin{figure*}[!t]
    \centering
    \centerline{\includegraphics[width=1\linewidth]{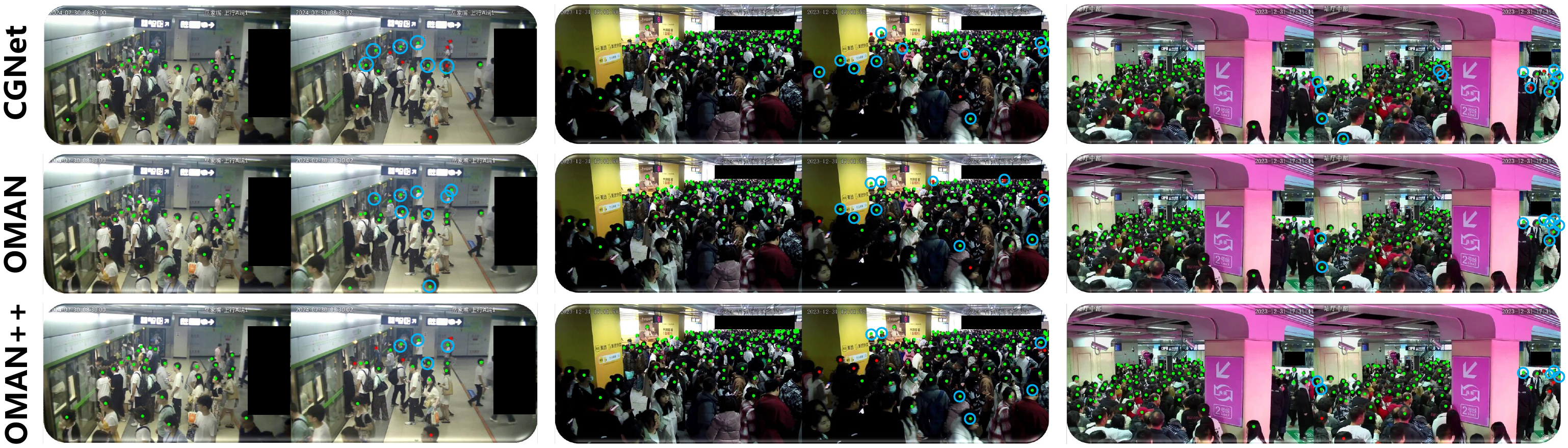}}
    \vspace{-5pt}
    \caption{\textbf{Visualization cases on \OurDataset dataset} In each image pair, green dots on the left sub-image represent location results of reference pedestrians in previous frame; while in the current frame on the right, green dots denotes 'pedestrian' class, red dots represents 'inflow' class, black rectangles are masked regions, and blue circles highlight the failure cases of matching according to the ground-truth labels. The locating results for CGNet are kept the same with the ones for OMAN++}
    \label{fig:vis}
\end{figure*}

\vspace{5pt}
\textbf{Results on WuhanMetroCrowd.} To better demonstrate the superiority of OMAN++ in crowded and congested scenes, we conduct a comparison on our proposed \OurDataset dataset. Due to the lack of identity annotation, we exclude two identity-based approaches (DRNet\cite{DRNet} and FMDC\cite{FMDC}) and the unreleased PDTR\cite{PDTR}, and we conduct comprehensive evaluations on \OurDataset dataset (shown in Table~\ref{tab:Metro}). Generally, OMAN++ demonstrates significant improvements over MDC~\cite{MDC}, achieving $47.5\%$, $63.6\%$ and $38.1\%$ reductions in MAE, MSE and WRAE, respectively. Compared to OMAN~\cite{OMAN}, OMAN++ still reduces MAE, MSE and WRAE by $35.9\%$, $43.6\%$ and $37.7\%$. Particularly, OMAN++ also achieves the lowest WRAE across five scenes, including platform, transfer, fare gate, lobby and escalator, and the second lowest WRAE in exit/entrance. We observe that all the approaches fail in Lobby with a WRAE even above $60\%$, but OMAN++ lower the WRAE by $31.4\%\sim53.8\%$. Considering that lobby is one of the most densest scenes for its largest capacity, OMAN++ indeed shows a great adaptability to crowds. 
Notably, despite the strong results, the remaining performance gap highlights both the challenging nature of our \OurDataset benchmark and opportunities for future improvement in VIC. To illustrate this, we provide some challenging cases in Fig.~\ref{fig:vis}.

\vspace{5pt}
\textbf{Efficiency Comparison.} OMAN++ demonstrates significant advantages over other methods across computational efficiency (Table~\ref{tab:efficiency}). Note that CGNet~\cite{CGNet} does not incorporate a locator. Despite higher parameters, our methods achieve higher inference speed beyond all density map-based methods through efficient resource utilization. OMAN++ processes frames in a coarse-to-fine manner and implements matching at the individual level rather than on the whole image, facilitating real-time employment in the real world.

\begin{table}[!t]
    \caption{\textbf{Efficiency of different methods}}
    \centering
    \label{tab:efficiency}
    \renewcommand{\arraystretch}{0.95} 
    \addtolength{\tabcolsep}{14 pt} 
    \begin{tabular}{@{}lcrc@{}}
        \toprule
        Method&Counter&FPS& \#Param. \\
        \midrule
        DRNet&Density Map&7.01&18.76M \\
        CGNet&- (Localization)&14.45&49.45M \\
        FMDC&Density Map&7.49&28.30M \\
        MDC&Density Map&2.46&31.46M \\
        \midrule
        OMAN&Localization&11.42&79.91M\\
        OMAN++&Localization&7.86&81.20M\\
        \bottomrule
    \end{tabular}
\end{table}

\subsection{Ablation Study}
\label{ssec:exp4}

\begin{table*}[!t]
    \label{ablation}
    \caption{\textbf{Ablation Study}}
    \vspace{-8pt}
    \begin{minipage}[t]{0.325\linewidth}
        \centering
        \subtable[\textbf{Impact of social grouping prior}. We evaluate each module based on social grouping prior without guidance of displacement
        \vspace{8pt}
        ]{
            \label{tab:ablation-methods-group}
            \addtolength{\tabcolsep}{-1.5 pt}
            \begin{tabular}{@{}lcccc@{}}
                \toprule
                ICG & OMPM & MAE$\downarrow$ & MSE$\downarrow$ & WRAE(\%)$\downarrow$ \\
                \midrule
                &&16.9&27.2&22.0 \\
                \Checkmark&&9.6&19.8&12.9 \\
                &\Checkmark&8.8&16.4&11.8 \\
                \Checkmark&\Checkmark&8.3&16.0&11.4 \\
                \bottomrule
            \end{tabular}}
    \end{minipage}
    \hfill
    \begin{minipage}[t]{0.325\linewidth}
        \centering
        \subtable[\textbf{Impact of spatial-temporal displacement prior WITHOUT $\mathcal{L}_{\tt D-OT}$}. We encapsulate displacement prior into the model via DASA and modulation]{
            \label{tab:ablation-methods-disp}
            \addtolength{\tabcolsep}{-3.5 pt}
            \begin{tabular}{@{}lcccc@{}}
                \toprule
                DASA & Modulator & MAE$\downarrow$ & MSE$\downarrow$ & WRAE(\%)$\downarrow$ \\
                \midrule                
                &&8.3&16.0&11.4 \\
                \Checkmark&&8.8&16.7&11.9 \\
                &\Checkmark&8.4&15.2&11.1 \\
                \Checkmark&\Checkmark&8.4&16.5&11.0 \\
                \bottomrule
            \end{tabular}}
    \end{minipage}
    \hfill
    \begin{minipage}[t]{0.325\linewidth}
        \centering
        \subtable[\textbf{Impact of spatial-temporal displacement prior WITH $\mathcal{L}_{\tt D-OT}$}. We gradually apply DASA and the modulator under the supervision of D-OT loss]{
            \label{tab:ablation-methods-disp2}
            \addtolength{\tabcolsep}{-3.5 pt}
            \begin{tabular}{@{}lcccc@{}}
                \toprule
                DASA & Modulator & MAE$\downarrow$ & MSE$\downarrow$ & WRAE(\%)$\downarrow$ \\
                \midrule
                &&8.3&16.2&11.2 \\
                \Checkmark&&8.2&16.0&11.0\\
                &\Checkmark&7.9&14.8&10.8 \\
                \Checkmark&\Checkmark&7.7&14.6&10.4 \\
                \bottomrule
            \end{tabular}}
    \end{minipage}
    
    \vspace{5pt}
    \hfill
    \begin{minipage}[t]{0.35\linewidth}
        \centering
        \subtable[\textbf{Fusion approach of displacement prior}]{
            \label{tab:ablation-fusion}
            \renewcommand{\arraystretch}{1} 
            \addtolength{\tabcolsep}{0 pt} 
            \begin{tabular}{@{}lccc@{}}
                \toprule
                Approach&MAE$\downarrow$&MSE$\downarrow$&WRAE(\%)$\downarrow$ \\
                \midrule
                Concatenation&8.5&16.6&10.8 \\
                Modulation&7.7&14.6&10.4 \\
                \bottomrule
            \end{tabular}}
        \subtable[\textbf{Impact of prior cost}]{
            \label{tab:ablation-disp}
            \renewcommand{\arraystretch}{1} 
            \addtolength{\tabcolsep}{0 pt} 
            \begin{tabular}{@{}lccc@{}}
                \toprule
                Approach&MAE$\downarrow$&MSE$\downarrow$&WRAE(\%)$\downarrow$ \\
                \midrule
                Displacement&8.6&17.1&10.9\\
                Prior Cost&7.7&14.6&10.4 \\
                \bottomrule
            \end{tabular}}
    \end{minipage}
    \hfill
    \begin{minipage}[t]{0.6\linewidth}
        \centering
        \subtable[\textbf{Impact of interval $\theta$ on SenseCrowd dataset.} D$0$$\sim$D$4$ represents different crowd densities, ranging [$0$,$50$), [$50$,$100$), [$100$,$150$), [$150$,$200$) and [$200$,$+\infty$). Best performance is in \textbf{boldface}, and second best is \underline{underlined}]{
            \label{tab:theta}
            \renewcommand{\arraystretch}{1} 
            \addtolength{\tabcolsep}{0 pt} 
                \begin{tabular}{@{}lcccccccc@{}}
                    \toprule
                    \multirow{2}{*}{$\theta$} & \multirow{2}{*}{MAE$\downarrow$} & \multirow{2}{*}{MSE$\downarrow$} & \multirow{2}{*}{WRAE(\%)$\downarrow$} & \multicolumn{5}{c}{Density\textbf{ }level (MAE$\downarrow$)} \\
                    \cmidrule{5-9}
                    & & & & D$0$ & D$1$ & D$2$ & D$3$ & D$4$ \\
                    \midrule
                    5 (1s) & 12.7 & 20.8 & 18.6 & 6.7 & 9.8 & 31.1 & 38.4 & \textbf{33.6} \\
                    10 (2s) & \textbf{7.6} & \textbf{13.5} & \underline{11.3} & 4.2 & \textbf{5.9} & 13.3 & \underline{15.6} & \underline{36.4} \\
                    15 (3s) & \underline{7.7} & \underline{14.6} & \textbf{10.4} & \textbf{3.3} & \underline{6.3} & \textbf{12.3} & \textbf{12.5} & 47.9 \\
                    20 (4s) & 10.5 & 19.1 & 13.8 & \underline{4.1} & 9.4 & \underline{12.6} & 21.3 & 68.5 \\
                    25 (5s) & 12.1 & 21.0 & 15.7 & 4.3 & 11.3 & 16.6 & 27.1 & 71.8 \\
                    \bottomrule
                \end{tabular}}
    \end{minipage}
    \vspace{-8pt}
\end{table*}

\textbf{Impact of Proposed Modules.} Here we ablate our approach to justify the contribution of the social grouping prior in Table~\ref{tab:ablation-methods-group} and of the displacement prior in Table~\ref{tab:ablation-methods-disp} and Table~\ref{tab:ablation-methods-disp2}. One can observe that, both the priors contribute to all the three main improvements. We first notice that precision dropped when we apply only DASA without the supervision of $\mathcal{L}_{\tt D-OT}$. We attribute it mainly to insufficient supervision of prior embedding $P$, due to $P$ is only indirectly supervised by appearance-based constraints. We also observe that, besides the constraints on the learning of the displacement prior, the prior cost can also guide the model to filter out some errors due to the slight improvement when only $\mathcal{L}_{\tt D-OT}$ is applied. Finally, the integration of both priors yields synergistic performance gains, demonstrating the effectiveness of our cohesive framework.


\vspace{5pt}
\textbf{Fusion Approach of Displacement Prior.} To evaluate the effectiveness of our proposed displacement-aware modulation for OMPM module, we compare the modulation with directly concatenating the displacement prior embeddings with appearance features. The comparison result in Table~\ref{tab:ablation-fusion} demonstrates that our coarse-to-fine modulation achieves a better balance between motion and appearance. It is mainly due to that
concatenation introduces feature redundancy and conflicts, while our modulation adaptively recalibrates the appearance features guided by the geometric prior, thereby enhancing the performance of O2M matching.

\begin{figure}[!t]
	\centering  
	\subfigbottomskip=3pt 
	\subfigcapskip=-5pt 
	\subfigure[\textbf{Performance of O2M head with different layer numbers.} The horizontal axis indicates the number of layers.]{
		\hspace*{-0.4em}\includegraphics[width=0.93\linewidth, height=0.45\linewidth]{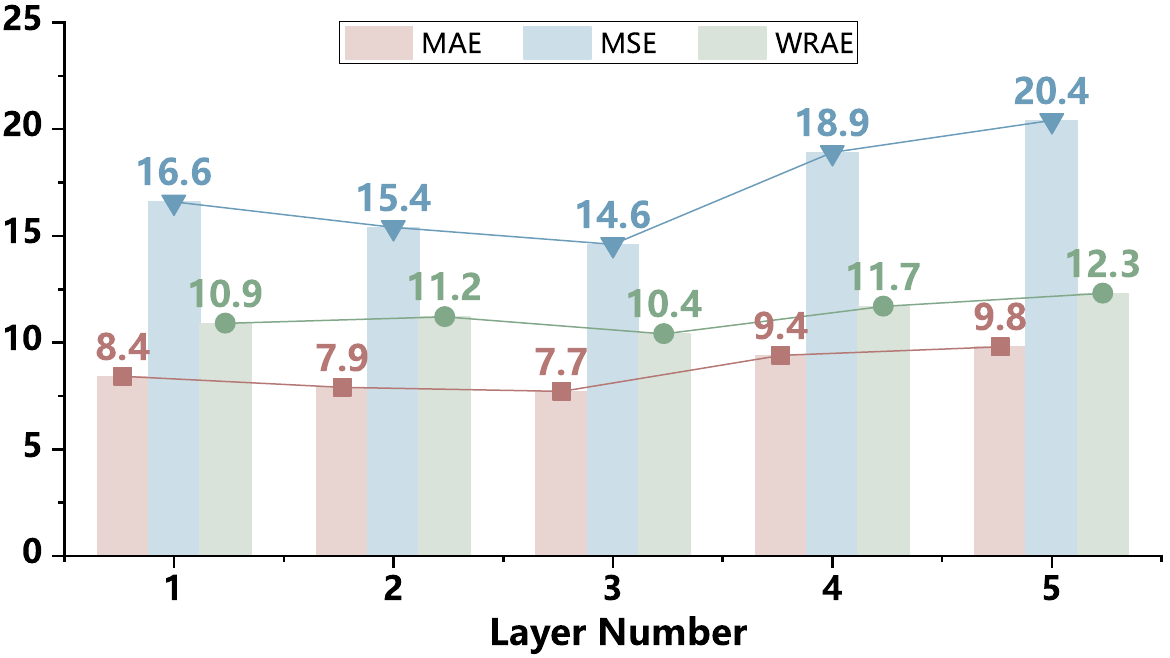}
            \label{fig:ablation-mlp}}
        \quad
	\subfigure[\textbf{Impact of $\lambda$.} The horizontal axis indicates the value of $\lambda$.]{
		\hspace*{-0.4em}\includegraphics[width=0.93\linewidth, height=0.45\linewidth]{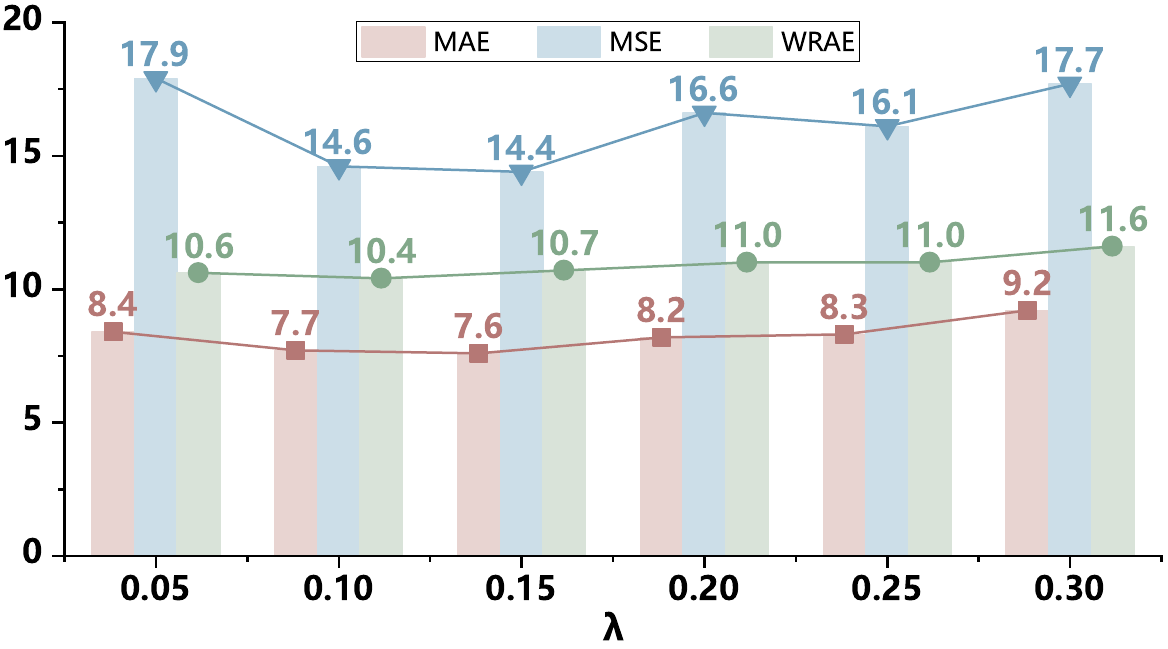}
            \label{fig:ablation-lambda}}
    \vspace{-5pt}
	\caption{\textbf{Impact of parameters.}}
    \label{fig:ablation}
\end{figure}

\vspace{5pt}
\textbf{Performance with Different Layer Numbers.} Since we implement our O2M head by a simple MLP, we conduct an experiment on the SenseCrowd dataset to determine the optimal layer configuration. As shown in Fig.~\ref{fig:ablation-mlp}, the performance is comparable with state-of-the-art methods with a shallow network, while too many layers can lead to overfitting, and the MAE, MSE, and WRAE all reach the minimum when the number of MLP layers is equal to three. 

\vspace{5pt}
\textbf{Impact of $\lambda$.} In the initial stage of training, the loss for appearance should take the dominance to establish a coarse matching, because motion can only be learned after preliminary associations are formed. 
Subsequently, the constraint of displacement should be gradually enhanced to provide a fine-grained correspondence. 
Empirically, we find that a constant $\lambda$ is sufficient because the appearance loss decays fast initially, allowing the displacement constraint to self-adjust its influence.
We therefore 
show the optimal $\lambda$ for OMAN++ is $0.1$ in Fig.~\ref{fig:ablation-lambda}. We also observe that setting $\lambda$ to $0.15$ enables OMAN++ to achieve better performance in dense scenes (lower MAE and MSE), which is consistent with our analysis above, but it slightly sacrifices the performance in sparser scenes (higher WRAE).

\subsection{Discussion}
\label{sec:disc}

\textbf{Analysis of prior cost.} 
To justify our learnable prior cost over static metrics like Euclidean distance, we first depict the relationship between prior cost and displacement in Fig.~\ref{fig:cost_wrt_disp}.
One can observe that prior cost tends to be more sensitive to displacement in x-axis than in y-axis, which simply verifies our grounds that OMAN++ models the anisotropic patterns of motion 
caused by camera projection, scene layout and pedestrian mobility that may induce axis-dependent statistics. Projecting the raw displacement prior into a learned latent space \romannumeral1) compensates for axis scaling and correlation; \romannumeral2) allows non-linear sensitivity to displacement magnitudes; and \romannumeral3) produces a displacement-aware term that fuses naturally with appearance similarity. 
To further demonstrate the effectiveness, we compare the results by replacing our prior cost with the displacement itself 
(Table~\ref{tab:ablation-disp}).

\vspace{3pt}
\textbf{Impact of frame interval $\theta$.} For a fair comparison, we simply keep the same frame interval $\theta$ as existing works; but for a comprehensive understanding of the $\theta$, we evaluate OMAN++ with different $\theta$ values in Table~\ref{tab:theta}. The model achieves the lowest WRAE when $\theta$ is set to $3$, consistent with CGNet~\cite{CGNet}.
We also observe that in densest scenarios (D$4$), performance improves with smaller intervals, likely because dense crowds exhibit minimal structural change over shorter periods, making motion-wise correlation easier to capture.

\section{Conclusion}
\label{sec:conc}

In this work, we first construct a novel VIC benchmark, \OurDataset, to fill current data gap of crowded scenarios. \OurDataset features diverse density levels, long video sequences, dynamic pedestrian flows, and intensive occlusion.
To further adapt VIC methods to dense crowds, we conclude our two observations into a social group prior and a spatial-temporal displacement prior. 
Based on the two informative priors, we introduce a simple but effective model called OMAN++ for VIC. We first implicitly generate token-to-token similarity context with an ICG module and implement O2M assignment by an OMPM module, to inform the group context. We then exploit the displacement prior with a DPI module, infusing the appearance and motion cues through a DASA block, a displacement-aware modulator, and a D-OT loss. 
Experiments on the three standard benchmarks and \OurDataset dataset indicate state-of-the-art performance. Due to the simplicity of implementation and a fresh perspective, we believe OMAN++ will serve as a strong baseline for VIC with ample scope for improvement.



\IEEEpubidadjcol

\section*{Acknowledgments}
The authors would like to thank FiberHome Telecommunication Technologies Co., Ltd. and Wuhan Metro Group Co., Ltd. for providing the data.

\bibliographystyle{IEEEtran}
\bibliography{refs}

@STRING{IEEE_J_TITS        = "{IEEE} Trans. Intell. Transp. Syst."}

@string(Naval_Research_Logistics_Quarterly = {Nav. Res. Logistics Quart.})

@String(TPAMI = {{IEEE} Trans. Pattern Anal. Mach. Intell.})

@String(IJCV = {Int. J. Comput. Vis.})

@String(CVPR = {{IEEE} Conf. Comput. Vis. Pattern Recognit.})

@String(ICCV = {{IEEE} Int. Conf. Comput. Vis.})

@String(ECCV = {Eur. Conf. Comput. Vis.})

@String(AAAI = {AAAI Conf. Artif. Intell.})

@String(ACMMM = {ACM Int. Conf. Multimedia})

@String(ICME = {Int. Conf. Multimedia and Expo})

@String(ICIP = {{IEEE} Int. Conf. Image Proc.})

@String(TIP = {{IEEE} Trans. Image Proc.})

@String(TCSVT = {{IEEE} Trans. Circuit Syst. Video Technol.})

@String(TIM = {{IEEE} Trans. Instrum. Meas.})

@String(WACV = {Winter Conf. Appl. Comput. Vis.})

@String(Neurocomputing = {Neurocomputing})

@INPROCEEDINGS{DRNet,
  author = {Han, Tao and Bai, Lei and Gao, Junyu and Wang, Qi and Ouyang, Wanli},
  booktitle = CVPR,
  title = {{DR.VIC}: Decomposition and Reasoning for Video Individual Counting},
  volume = {},
  number = {},
  pages = {3073-3082},
  year = {2022},
  doi = {10.1109/CVPR52688.2022.00309},
}

@INPROCEEDINGS{CGNet,
  author = {Liu, Xinyan and Li, Guorong and Qi, Yuankai and Yan, Ziheng and Han, Zhenjun and van den Hengel, Anton and Yang, Ming-Hsuan and Huang, Qingming},
  booktitle = CVPR, 
  title = {Weakly Supervised Video Individual Counting}, 
  volume = {},
  number = {},
  pages = {19228-19237},
  year = {2024},
  doi = {10.1109/CVPR52733.2024.01819},
}

@INPROCEEDINGS{FMDC,
  author={Wan, Chang-Lin and Huang, Feng-Kai and Shuai, Hong-Han},
  booktitle = WACV, 
  title={Density-Based Flow Mask Integration via Deformable Convolution for Video People Flux Estimation}, 
  year={2024},
  volume={},
  number={},
  pages={6559-6568},
  doi={10.1109/WACV57701.2024.00644},
}

@INPROCEEDINGS{PDTR,
  author = {Li, Rui and Liu, Yishu and Li, Huafeng and Li, Jinxing and Lu, Guangming},
  booktitle = ACMMM,
  title = {Prototype-Guided Dual-Transformer Reasoning for Video Individual Counting},
  year = {2024},
  volume={},
  number={},
  pages = {10258–10267},
  doi = {10.1145/3664647.3681211},
}

@INPROCEEDINGS{CACViT,
  author = {Wang, Zhicheng and Xiao, Liwen and Cao, Zhiguo and Lu, Hao},
  booktitle = AAAI,
  title = {Vision transformer off-the-shelf: a surprising baseline for few-shot class-agnostic counting},
  year = {2024},
  volume={},
  number={},
  pages = {5832–5840},
  doi = {10.1609/aaai.v38i6.28396},
}

@ARTICLE{LOI-tip,
  author={Chan, Antoni B. and Vasconcelos, Nuno},
  journal=TIP, 
  title={Counting People With Low-Level Features and Bayesian Regression}, 
  year={2012},
  volume={21},
  number={4},
  pages={2160-2177},
  doi={10.1109/TIP.2011.2172800}}

@INPROCEEDINGS{LOI-2013,
  author = {Ma, Zheng and Chan, Antoni B.},
  booktitle = CVPR,
  title = {Crossing the Line: Crowd Counting by Integer Programming with Local Features},
  year = {2013},
  volume={},
  number={},
  pages = {2539-2546},
  doi = {10.1109/CVPR.2013.328},
}

@INPROCEEDINGS{LOI-2016,
  author = {Zhao, Zhuoyi and Li, Hongsheng and Zhao, Rui and Wang, Xiaogang},
  booktitle = ECCV,
  title = {Crossing-Line Crowd Counting with Two-Phase Deep Neural Networks},
  year = {2016},
  volume={},
  number={},
  pages = {712-726},
  doi = {10.1007/978-3-319-46484-8_43},
}

@ARTICLE{LOI-2019,
  author = {Zheng, Huicheng and Lin, Zijian and Cen, Jiepeng and Wu, Zeyu and Zhao, Yadan},
  journal = TCSVT,
  title = {Cross-Line Pedestrian Counting Based on Spatially-Consistent Two-Stage Local Crowd Density Estimation and Accumulation},
  year = {2019},
  volume={29},
  number={3},
  pages = {787-799},
  doi = {10.1109/TCSVT.2018.2807806},
}

@INPROCEEDINGS{HeadHunter-T,
  author = {Sundararaman, Ramana and De Almeida Braga, Cédric and Marchand, Eric and Pettré, Julien},
  booktitle = CVPR,
  title = {Tracking Pedestrian Heads in Dense Crowd},
  year = {2021},
  volume={},
  number={},
  pages = {3864-3874},
  doi = {10.1109/CVPR46437.2021.00386},
}

@ARTICLE{FairMOT,
  author = {Zhang, Yifu and Wang, Chunyu and Wang, Xinggang and Zeng, Wenjun and Liu, Wenyu},
  journal = IJCV,
  title = {{FairMOT}: On the Fairness of Detection and Re-identification in Multiple Object Tracking},
  year = {2021},
  volume={129},
  number={11},
  pages = {3069–3087},
  doi = {10.1007/s11263-021-01513-4},
}

@INPROCEEDINGS{ByteTrack,
  author = {Zhang, Yifu and Sun, Peize and Jiang, Yi and Yu, Dongdong and Weng, Fucheng and Yuan, Zehuan and Luo, Ping and Liu, Wenyu and Wang, Xinggang},
  booktitle = ECCV,
  title = {{ByteTrack}: Multi-object Tracking by Associating Every Detection Box},
  year = {2022},
  volume={},
  number={},
  pages = {1–21},
  doi = {10.1007/978-3-031-20047-2_1},
}

@ARTICLE{Hungarian,
  author = {Kuhn, Harold W},
  journal = Naval_Research_Logistics_Quarterly,
  title = {The Hungarian method for the assignment problem},
  year = {1955},
  volume={2},
  number={1-2},
  pages = {83-97},
  doi = {10.1002/nav.3800020109},
}

@INPROCEEDINGS{PET,
  author = {Liu, Chengxin and Lu, Hao and Cao, Zhiguo and Liu, Tongliang},
  booktitle = ICCV,
  title = {Point-Query Quadtree for Crowd Counting, Localization, and More},
  year = {2023},
  volume={},
  number={},
  pages = {1676-1685},
  doi = {10.1109/ICCV51070.2023.00161},
}

@INPROCEEDINGS{LSTN,
  author = {Fang, Yanyan and Zhan, Biyun and Cai, Wandi and Gao, Shenghua and Hu, Bo},
  booktitle = ICME,
  title = {Locality-Constrained Spatial Transformer Network for Video Crowd Counting},
  year = {2019},
  volume={},
  number={},
  pages = {814-819},
  doi = {10.1109/ICME.2019.00145},
}

@ARTICLE{FTAN,
  author = {Wu, Xingjiao and Xu, Baohan and Zheng, Yingbin and Ye, Hao and Yang, Jing and He, Liang},
  journal = Neurocomputing,
  title = {Fast video crowd counting with a Temporal Aware Network},
  year = {2020},
  volume={403},
  number={},
  pages = {13-20},
  doi = {10.1016/j.neucom.2020.04.071},
}

@INPROCEEDINGS{TransMOT,
  author = {Chu, Peng and Wang, Jiang and You, Quanzeng and Ling, Haibin and Liu, Zicheng},
  booktitle = WACV,
  title = {{TransMOT}: Spatial-Temporal Graph Transformer for Multiple Object Tracking},
  year = {2023},
  volume={},
  number={},
  pages = {4859-4869},
  doi = {10.1109/WACV56688.2023.00485},
}

@ARTICLE{Group-reid,
  author = {Yan, Yichao and Qin, Jie and Ni, Bingbing and Chen, Jiaxin and Liu, Li and Zhu, Fan and Zheng, Wei-Shi and Yang, Xiaokang and Shao, Ling},
  journal = TPAMI,
  title = {Learning Multi-Attention Context Graph for Group-Based Re-Identification},
  year = {2023},
  volume={45},
  number={6},
  pages = {7001-7018},
  doi = {10.1109/TPAMI.2020.3032542},
}

@ARTICLE{SenseCrowd,
  author = {Li, Haopeng and Liu, Lingbo and Yang, Kunlin and Liu, Shinan and Gao, Junyu and Zhao, Bin and Zhang, Rui and Hou, Jun},
  journal = TIP,
  title = {Video Crowd Localization With Multifocus Gaussian Neighborhood Attention and a Large-Scale Benchmark},
  year = {2022},
  volume={31},
  number={},
  pages = {6032–6047},
  doi = {10.1109/TIP.2022.3205210},
}

@article{OT,
author = {Peyr{\'e}, Gabriel and Cuturi, Marco and others},
title = {Computational Optimal Transport: With Applications to Data Science},
year = {2019},
issue_date = {Feb. 2019},
publisher = {Now Publishers Inc.},
address = {Hanover, MA, USA},
volume = {11},
number = {5–6},
issn = {1935-8237},
doi = {10.1561/2200000073},
journal = {Found. Trends Mach. Learn.},
pages = {355–607},
numpages = {257}
}

@INPROCEEDINGS{LOI-2009,
  author = {Cong, Yang and Gong, Haifeng and Zhu, Song-Chun and Tang, Yandong},
  booktitle = CVPR,
  title = {Flow mosaicking: Real-time pedestrian counting without scene-specific learning},
  year = {2009},
  volume={},
  number={},
  pages = {1093-1100},
  doi = {10.1109/CVPR.2009.5206648},
}

@INPROCEEDINGS{LOI-track1,
  author = {Chen, Tsong-Yi and Chen, Chao-Ho and Wang, Da-Jinn and Kuo, Yi-Li},
  booktitle = {Int. Conf. Genetic Evol. Comput.},
  title = {A People Counting System Based on Face-Detection},
  year = {2010},
  volume={},
  number={},
  pages = {699-702},
  doi = {10.1109/ICGEC.2010.178},
}

@article{LOI-track4,
author = {Dan, Byoung-Kyu and Kim, You-Sun and Suryanto and Jung, June-Young and Ko, Sung-Jea},
title = {Robust people counting system based on sensor fusion},
year = {2012},
issue_date = {Aug. 2012},
volume = {58},
number = {3},
doi = {10.1109/TCE.2012.6311350},
journal = {IEEE Trans. Consum. Electron.},
pages={1013--1021},
numpages = {8}
}

@article{LOI-2016-tcsvt,
author = {Ma, Zheng and Chan, Antoni B.},
title = {Counting People Crossing a Line Using Integer Programming and Local Features},
year = {2016},
issue_date = {Oct. 2016},
volume = {26},
number = {10},
doi = {10.1109/TCSVT.2015.2489418},
journal = TCSVT,
pages = {1955-1969},
numpages = {14}
}

@INPROCEEDINGS{VCC-det1,
  author={Idrees, Haroon and Saleemi, Imran and Seibert, Cody and Shah, Mubarak},
  booktitle=CVPR, 
  title={Multi-source Multi-scale Counting in Extremely Dense Crowd Images}, 
  year={2013},
  volume={},
  number={},
  pages={2547-2554},
  doi={10.1109/CVPR.2013.329}}

@INPROCEEDINGS{VCC-det2,
  author={Subburaman, Venkatesh Bala and Descamps, Adrien and Carincotte, Cyril},
  booktitle={Int. Conf. Adv. Video  and Signal-Based Surveillance}, 
  title={Counting People in the Crowd Using a Generic Head Detector}, 
  year={2012},
  volume={},
  number={},
  pages={470-475},
  doi={10.1109/AVSS.2012.87}}

@ARTICLE{VCC-det3,
  author={Sam, Deepak Babu and Peri, Skand Vishwanath and Sundararaman, Mukuntha Narayanan and Kamath, Amogh and Babu, R. Venkatesh},
  journal=TPAMI, 
  title={Locate, Size, and Count: Accurately Resolving People in Dense Crowds via Detection}, 
  year={2021},
  volume={43},
  number={8},
  pages={2739-2751},
  doi={10.1109/TPAMI.2020.2974830}}

@INPROCEEDINGS{VCC-det4,
  author={Liu, Yuting and Shi, Miaojing and Zhao, Qijun and Wang, Xiaofang},
  booktitle=CVPR, 
  title={Point in, Box Out: Beyond Counting Persons in Crowds}, 
  year={2019},
  volume={},
  number={},
  pages={6462-6471},
  doi={10.1109/CVPR.2019.00663}}

@article{VCC-dens1,
  title={Learning to count objects in images},
  author={Lempitsky, Victor and Zisserman, Andrew},
  journal={Adv. in neural Inf. Process. Syst.},
  volume={23},
  year={2010}
}

@INPROCEEDINGS{VCC-dens4,
  author={Liu, Weizhe and Salzmann, Mathieu and Fua, Pascal},
  booktitle=CVPR, 
  title={Context-Aware Crowd Counting}, 
  year={2019},
  volume={},
  number={},
  pages={5094-5103},
  doi={10.1109/CVPR.2019.00524}}

@INPROCEEDINGS{VCC-loc1,
  author={Song, Qingyu and Wang, Changan and Jiang, Zhengkai and Wang, Yabiao and Tai, Ying and Wang, Chengjie and Li, Jilin and Huang, Feiyue and Wu, Yang},
  booktitle=ICCV, 
  title={Rethinking Counting and Localization in Crowds: A Purely Point-Based Framework}, 
  year={2021},
  volume={},
  number={},
  pages={3345-3354},
  doi={10.1109/ICCV48922.2021.00335}}

@ARTICLE{VCC-loc2,
  author={Cheng, Jian and Xiong, Haipeng and Cao, Zhiguo and Lu, Hao},
  journal=TIP, 
  title={Decoupled Two-Stage Crowd Counting and Beyond}, 
  year={2021},
  volume={30},
  number={},
  pages={2862-2875},
  doi={10.1109/TIP.2021.3055631}}

@INPROCEEDINGS{OCsort,
  author={Cao, Jinkun and Pang, Jiangmiao and Weng, Xinshuo and Khirodkar, Rawal and Kitani, Kris},
  booktitle=CVPR, 
  title={Observation-Centric SORT: Rethinking SORT for Robust Multi-Object Tracking}, 
  year={2023},
  volume={},
  number={},
  pages={9686-9696},
  doi={10.1109/CVPR52729.2023.00934}}

@article{Bot-sort,
  title={BoT-SORT: Robust Associations Multi-Pedestrian Tracking},
  author={Aharon, Nir and Orfaig, Roy and Bobrovsky, Ben-Zion},
  journal = {},
  note={arXiv:2206.14651},
  year={2022}
}

@inproceedings{JDE,
  title={Towards real-time multi-object tracking},
  author={Wang, Zhongdao and Zheng, Liang and Liu, Yixuan and Li, Yali and Wang, Shengjin},
  booktitle=ECCV,
  pages={107--122},
  year={2020},
  organization={Springer}
}

@inproceedings{chainedtracker,
  title={Chained-tracker: Chaining paired attentive regression results for end-to-end joint multiple-object detection and tracking},
  author={Peng, Jinlong and Wang, Changan and Wan, Fangbin and Wu, Yang and Wang, Yabiao and Tai, Ying and Wang, Chengjie and Li, Jilin and Huang, Feiyue and Fu, Yanwei},
  booktitle=ECCV,
  pages={145--161},
  year={2020},
  organization={Springer}
}

@inproceedings{centertrack,
  title={Tracking objects as points},
  author={Zhou, Xingyi and Koltun, Vladlen and Kr{\"a}henb{\"u}hl, Philipp},
  booktitle=ECCV,
  pages={474--490},
  year={2020},
  organization={Springer}
}

@article{apptracker+,
  title={Apptracker+: Displacement uncertainty for occlusion handling in low-frame-rate multiple object tracking},
  author={Zhou, Tao and Ye, Qi and Luo, Wenhan and Ran, Haizhou and Shi, Zhiguo and Chen, Jiming},
  journal=IJCV,
  pages={1--26},
  year={2024},
  publisher={Springer}
}

@inproceedings{trackformer,
  title={Trackformer: Multi-object tracking with transformers},
  author={Meinhardt, Tim and Kirillov, Alexander and Leal-Taixe, Laura and Feichtenhofer, Christoph},
  booktitle=CVPR,
  pages={8844--8854},
  year={2022}
}

@article{transtrack,
  title={Transtrack: Multiple object tracking with transformer},
  author={Sun, Peize and Cao, Jinkun and Jiang, Yi and Zhang, Rufeng and Xie, Enze and Yuan, Zehuan and Wang, Changhu and Luo, Ping},
  journal = {},
  note={arXiv:2012.15460},
  year={2020}
}

@INPROCEEDINGS{DFormerv2,
  author={Yin, Bo-Wen and Cao, Jiao-Long and Cheng, Ming-Ming and Hou, Qibin},
  booktitle=CVPR, 
  title={DFormerv2: Geometry Self-Attention for RGBD Semantic Segmentation}, 
  year={2025},
  volume={},
  number={},
  pages={19345-19355},
  doi={10.1109/CVPR52734.2025.01802}}

@article{sinkhorn,
  title={Concerning nonnegative matrices and doubly stochastic matrices},
  author={Richard Sinkhorn and Paul Knopp},
  journal={Pacific J. of Math.},
  year={1967},
  volume={21},
  pages={343-348},
}

@INPROCEEDINGS{OMAN,
  title={Video Individual Counting With Implicit One-to-Many Matching},
  author={Zhu, Xuhui and Xu, Jing and Wang, Bingjie and Dai, Huikang and Lu, Hao},
  booktitle=ICIP,
  year={2025}
}

@inproceedings{convnext,
  title={A convnet for the 2020s},
  author={Liu, Zhuang and Mao, Hanzi and Wu, Chao-Yuan and Feichtenhofer, Christoph and Darrell, Trevor and Xie, Saining},
  booktitle=CVPR,
  pages={11976--11986},
  year={2022}
}

@inproceedings{MDC,
  title={Video Individual Counting for Moving Drones},
  author={Fan, Yaowu and Wan, Jia and Han, Tao and Chan, Antoni B and Ma, Andy J},
  booktitle=ICCV,
  year={2025}
}

@ARTICLE{metro_sys,
  author={Liu, Lingbo and Chen, Jingwen and Wu, Hefeng and Zhen, Jiajie and Li, Guanbin and Lin, Liang},
  journal=IEEE_J_TITS, 
  title={Physical-Virtual Collaboration Modeling for Intra- and Inter-Station Metro Ridership Prediction}, 
  year={2022},
  volume={23},
  number={4},
  pages={3377-3391},
  doi={10.1109/TITS.2020.3036057}}

@ARTICLE{track-tip,
  author={Shaohua Kevin Zhou and Chellappa, R. and Moghaddam, B.},
  journal=TIP, 
  title={Visual tracking and recognition using appearance-adaptive models in particle filters}, 
  year={2004},
  volume={13},
  number={11},
  pages={1491-1506},
  doi={10.1109/TIP.2004.836152}}


 


\vspace{11pt}


\vfill

\end{document}